\documentclass[twoside,11pt]{article}

\usepackage{jmlr2e}

\ShortHeadings{Learning in Normal Graphs}{Francesco A. N. Palmieri}
\firstpageno{1}

\begin{document}

\title{A Comparison of Algorithms for Learning Hidden Variables in Normal Graphs}

\author{\name Francesco A. N. Palmieri \email francesco.palmieri@unina2.it \\
       \addr Dipartimento di Ingegneria Industriale e dell'Informazione\\
       Seconda Unversit\`a di Napoli (SUN) \\
       via Roma 29, Aversa (CE), Italy}

\editor{arXiv}

\maketitle

\begin{abstract}
A Bayesian factor graph reduced to normal form \citep{Forney2001} consists in  the interconnection of {\em diverter units} (or equal constraint units) and Single-Input/Single-Output (SISO) blocks. In this framework localized adaptation rules are explicitly derived from a constrained maximum likelihood (ML) formulation and from a minimum KL-divergence criterion using  KKT conditions. The learning algorithms  are compared  with two other updating equations based on a Viterbi-like and on a variational approximation respectively.  The performance of the various  algorithm is  verified on synthetic data sets for various architectures. The objective of this paper is to provide the programmer with explicit algorithms for rapid deployment of Bayesian graphs in the applications.  
\end{abstract}

\smallskip


\section{Introduction}

Graphical models are a collection of very promising paradigms for signal processing, adaptive control, artificial intelligence and theoretical physics. Since early proposals 
\citep{Pearl1988} the literature has seen the emergence of a considerable interest in this framework \citep{SPMag2010}(and the special issue) as it promises a considerable impact on the applications.  The opportunity of manipulating information bi-directionally and fusing very heterogeneous data into a unique framework  with limited supervision and strong adaptivity, has opened the way to new exciting opportunities for innovation.  

A review on graphical models is certainly not the purpose of this paper for which  we refer to the vast current literature (for introductory textbooks see for example \citep{Jordan1998} \citep{Neapolitan2004}, \citep{Korb2004}, \citep{Bishop2006} and \citep{Koller2010}). 

Bayesian graphs are presented as directed or undirected graphs, but a very appealing approach to visualization and manipulation is the Factor Graph (FG) representation, in the so-called {\em normal form} (FGn)  \citep{Forney2001} \citep{Loeliger2004}. Factor graphs assign variables to edges and functions to nodes, but  in a normal graph, through the use of {\em diverter units} (equal constraints), the graph is reduced to an architecture in which each pair of blocks is connected only by one directed edge. An FGn resembles a circuit diagram where belief messages are easily visualized as they enter and exit the various SISO factor-blocks.  Diverters act like {\em buses} and can be used as expansion nodes when we need to augment an existing model with new variables. The graph can become modular  and allow different sensor modalities, or new sources, to be easily fused into the same system. Parameter learning in this representation can be approached  in a simple a unified way because we can concentrate on a unique rule for training any factor-block in the system regardless of its location (visible or hidden).     

The topic of parameter learning in Bayesian networks  has been presented in the literature from various points of view  \citep{Binder1997} \citep{Heckerman1996}\citep{Ghahramani2004}\citep{Beal_Gha2004}
\citep{Bengio2009} with the EM approach being  the most popular reference paradigm.  Various updating equations have been proposed in reference to  specific architectures, such as  HMM \citep{Rabiner1989} \citep{Welch2003}, trees \citep{Zhang2004} \citep{Harmeling2011}, etc.  When the variables are instantiated (sharp messages), learning the conditional probability matrices simply reduces to co-occurrence counting. Conversely when the variables are hidden, or when the information is available via smooth distributions, the problem  becomes harder because we need to define  explicitly the cost function, and derive the updating equations. 

Our objective in this paper is to focus on the specifics of the learning algorithms 
as they may be embedded in arbitrary graphs that have both visible and hidden variables. The constraint is that the algorithms are localized and are the same for every branch in the graph, regardless of the graph topology and the edge location. Each single block can ``see" only incoming forward and backward belief messages and it must be able to adapt its conditional probability matrix solely on this local information. 

In these notes we re-work explicitly the learning problem with specific attention to confining the updating equations  inside the  SISO block of any factor graph architecture in normal form. We show how to derive a robust maximum likelihood 
(ML)  and a minimum divergence (KL) algorithm posing the problem as   constrained optimization problems and using the Karhush-Kuhn-Tucker (KKT) conditions \citep{Avriel2003}. The ML and the KL algorithms are compared to other  updates derived from different criteria.     

We focus on four different algorithms with the objective of providing a support to  the programmer that has to write his/her own software implementation. The four algorithms reviewed here seem to capture capture most of the approaches presented in the literature for parameter learning in Bayesian networks.  We compare the performances of the four algorithms on synthetic data to provide background experimental evidence.
To our knowledge no comparison of this sort  has ever been presented in the literature and the derivations have never been presented in a unified way. Various algorithms have been presented  with reference to specific graph topologies and sometimes it is left to the programmer  choose the specifics of the updates that may become  implementation-dependent. The normal graph formulation allows us to focus on algorithms that are explicit and that are applicable to any graph topology.   

We confine our attention to fixed architectures and to cycle-free graphs. 
The issues of learning the graph architecture \citep{Koller2010} \citep{Anandkumar2011} \citep{Harmeling2011} and that of dealing with loopy graphs
\citep{Yedidia2005} are important topic of current research and are left out from the following discussion. More work for parameter  learning in these frameworks will be addresses elsewhere. 

We also do not include gradient methods \citep{Koller2010} because they require propagation of derivatives that are not exactly localized. We also leave out all stochastic approaches, such as Gibbs' sampling methods, as they would compare poorly with the presented algorithms in terms  of convergence time.     

In Section \ref{sec:not} we review notations and the basics of belief propagation in normal graphs with some examples. We then address the parameter learning problem with three algorithms in Section \ref{sec:parl}. The first is based on a localized likelihood function, the second on a Kullback-Leibler divergence cost function and a third on a Viterbi-like approximation.  A Bayesian approach to learning is also presented with a variational approximation in Section \ref{sec:bay}.
In Section  \ref{sec:single} we provide detailed results of simulations on a single SISO block. In Section \ref{sec:larger}  we report simulations on two different graph configurations: a tree with a single latent variable and a deeper architecture with four layers. Some details and algorithmic derivations are confined Appendices A and B.  Conclusions and suggestions for future work are  included in Section \ref{sec:concl}

\section{Notations and Basics}
\label{sec:not}

To define notations and terminology, we briefly review the essentials of belief propagation in normal graphs. In our notation we avoid the upper arrows  \citep{Loeliger2004} and assign a direction to each variable branch for unambiguous definition of forward and backward messages. 
\begin{figure}
\includegraphics[width=14.0cm]{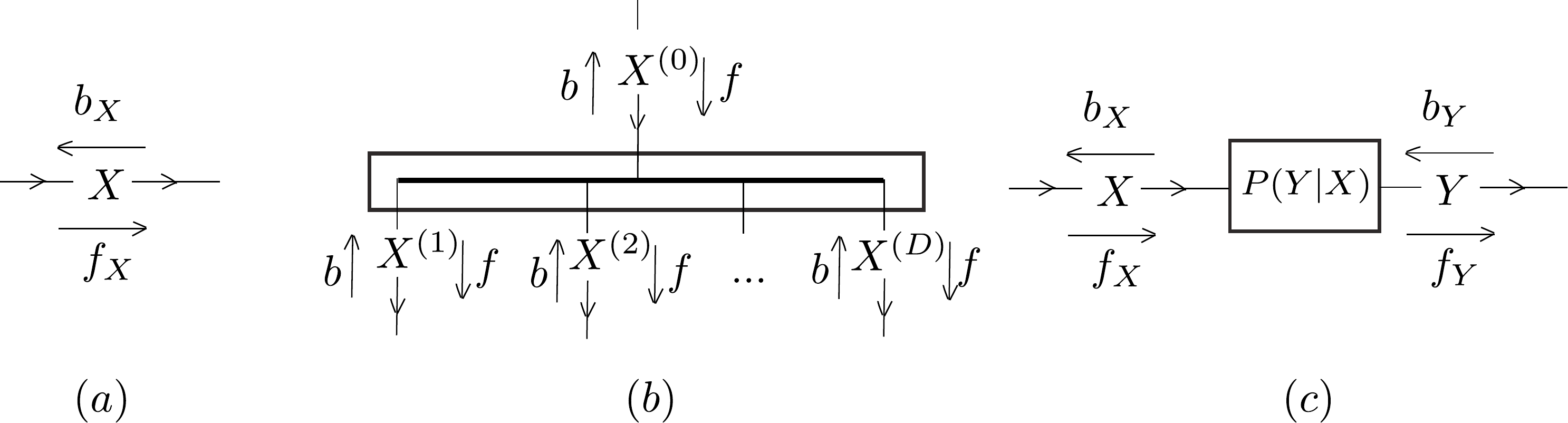}
\caption{(a) a variable branch; (b) a diverter; (c) a single block (factor).}
\label{fig:intro}
\end{figure}
\noindent
For a variable $X$ (Figure \ref{fig:intro}(a)) that takes values in the discrete alphabet ${\cal X}=\{x^1,x^2,...,x^{M_X} \}$, forward and backward messages are in function form $b_X(x^i)$ and $f_X(x^i)$, $i=1:M_X$ and in vector form ${\bf b}_X=(b_X(x^1), b_X(x^2),...,b_X(x^{M_X}))^T$ and ${\bf f}_X=(f_X(x^1), f_X(x^2),...,f_X(x^{M_X}))^T$.  All messages are proportional ($\propto$) to discrete distributions and  may be normalized to sum to one. Comprehensive knowledge about $X$ is contained in the distribution $p_X$ obtained through the product rule, $p_X(x^i) \propto f_X(x^i) b_X(x^i)$,
 $i=1:M_X$, in function form, or   ${\bf p}_X \propto {\bf f}_X \odot {\bf b}_X$, in vector form, where $\odot$ denotes the element-by-element product.

The {\em diverter} block (Figure \ref{fig:intro}(b)) represents the equality contraint with the variable $X$ replicated $D+1$ times.  Messages for incoming and outgoing branches carry different forward and backward information. Messages that leave the  block are obtained as the product of the incoming ones:
$b_{X^{(0)}}(x^i) \propto \prod_{j=1}^{D}  b_{X^{(j)}}(x^i)$; $f_{X^{(m)}} \propto f_{X^{(0)}}(x^i) \prod_{j=1, j \neq m}^{D}  b_{X^{(j)}}(x^i)$, $m=1:D$, $i=1:M_X$ in function form. 
In vector form: ${\bf b}_{X^{(0)}} \propto \odot_{j=1}^{D}  {\bf b}_{X^{(j)}}$; ${\bf f}_{X^{(m)}} \propto {\bf f}_{X^{(0)}} \odot_{j=1, j \neq m}^{D}  {\bf b}_{X^{(j)}}$, $m=1:D$.

The SISO block (Figure \ref{fig:intro}(c)) represents the conditional probability matrix of $X$ given $Y$. More specifically if $X$ takes values in the discrete alphabet ${\cal X}=\{x^1,x^2,...,x^{M_X} \}$ and $Y$ in  ${\cal Y}=\{y^1,y^2,...,y^{M_Y} \}$, $P(Y|X)$ is the 
$M_X \times M_Y$ row-stochastic matrix $P(Y|X)=[Pr\{ Y=y^j | X=x^i\}]_{i=1:M_X}^{j=1:M_Y}=[\theta_{ij}]_{i=1:M_X}^{j=1:M_Y}=\theta$. Outgoing messages are:  $f_Y(y^j) \propto \sum_{i=1}^{M_X} \theta_{ij} f_X(x^i)$;  $b_X(x^i) \propto \sum_{j=1}^{M_Y} \theta_{ij} b_Y(y^j)$, in function form. In vector form: ${\bf f}_Y \propto P(Y|X)^T {\bf f}_X$;  ${\bf b}_X \propto P(Y|X) {\bf b}_Y$. 
   
For the reader unfamiliar with this framework it should be emphasezed that the above rules are rigorous translation of Bayes' theorem and marginalization. It would be outside the scope of this paper to review the proofs for which we refer  to the classical papers \citep{Loeliger2004} \citep{Kschischang2001}.

 \begin{figure}[ht]
\begin{center}
\includegraphics[width=7.0cm]{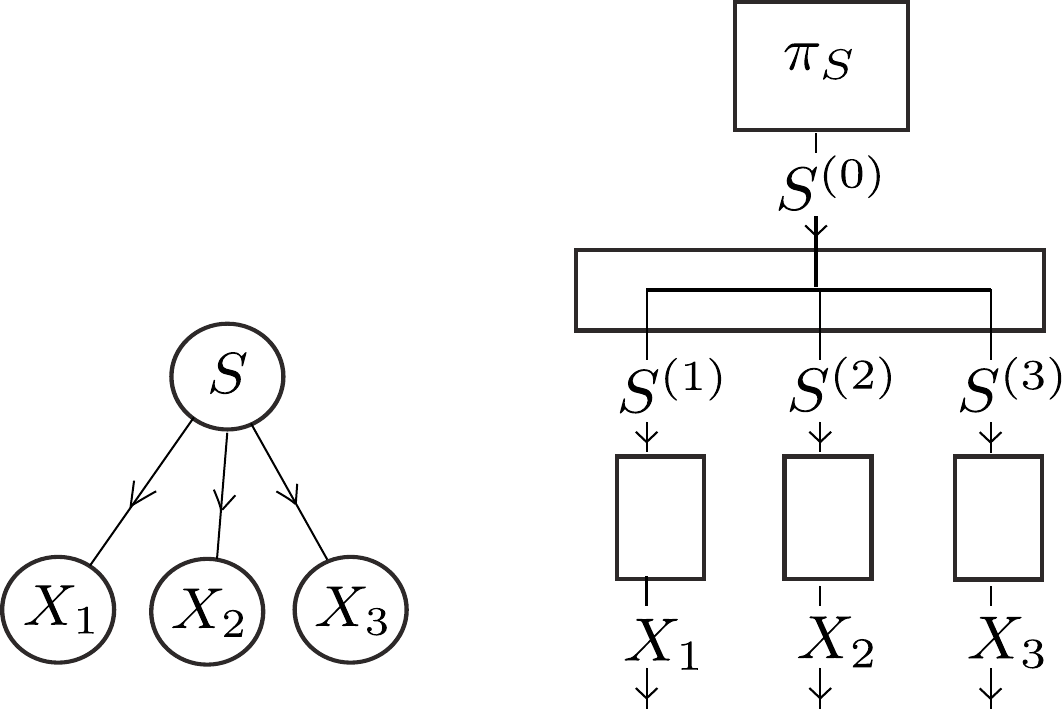}
\caption{Bayesian graph and factor graph in normal form for a single variable with three children.}
\end{center}
\label{fig:exx1}
\end{figure}

\begin{figure}[ht]
 \includegraphics[width=13.0cm]{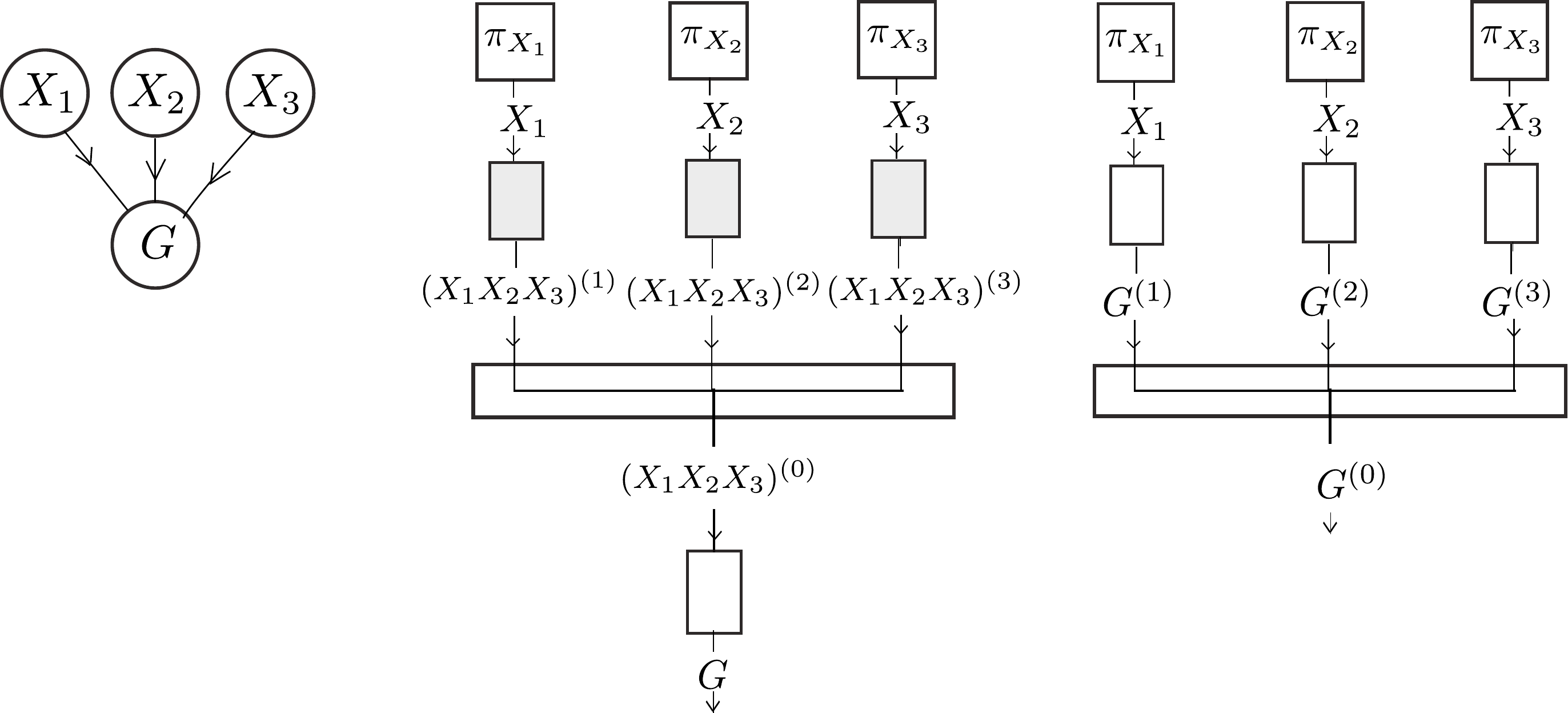}
\caption{Bayesian graph for three parents with a single child (left);  normal-form factor graph with the inclusion of the product space variable (middle); normal-form factor graph for a simplified model with direct connection to the child variable (right).}
\label{fig:exx2}
\end{figure}

\begin{figure}[ht]
\includegraphics[width=12.0cm]{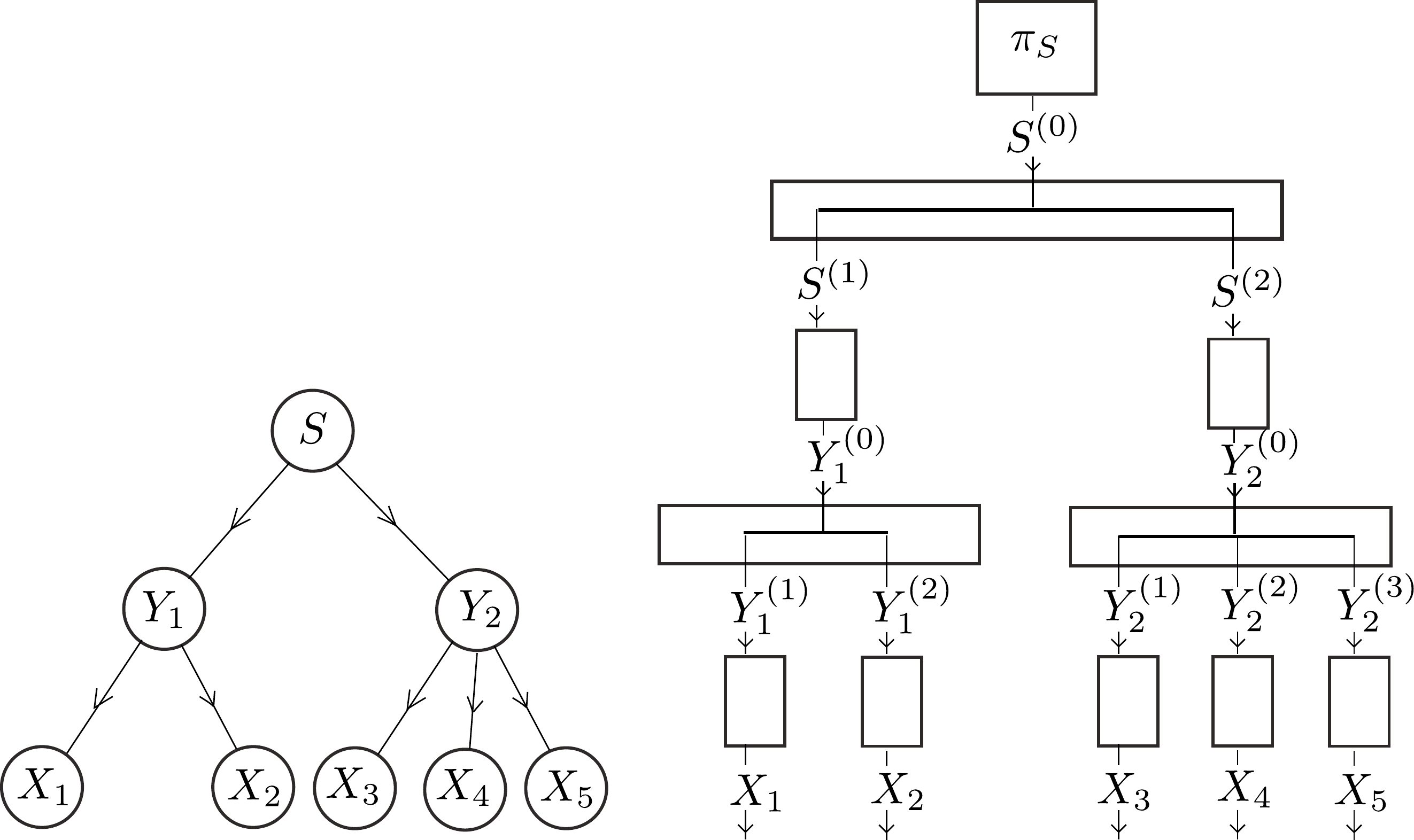}
\caption{The Bayesian graph and the FGn for a two-level hierarchy}
\label{fig:exx3}
\end{figure}

\begin{figure}[ht]
\includegraphics[width=5.0cm]{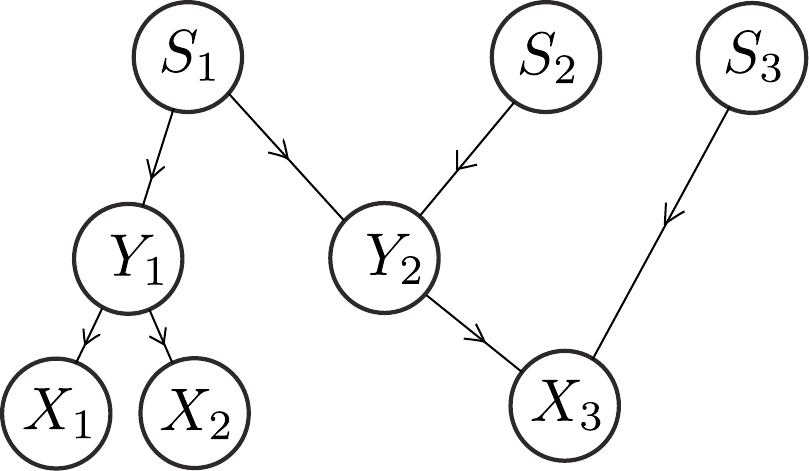} \vspace{0 cm}
\includegraphics[width=8.0cm]{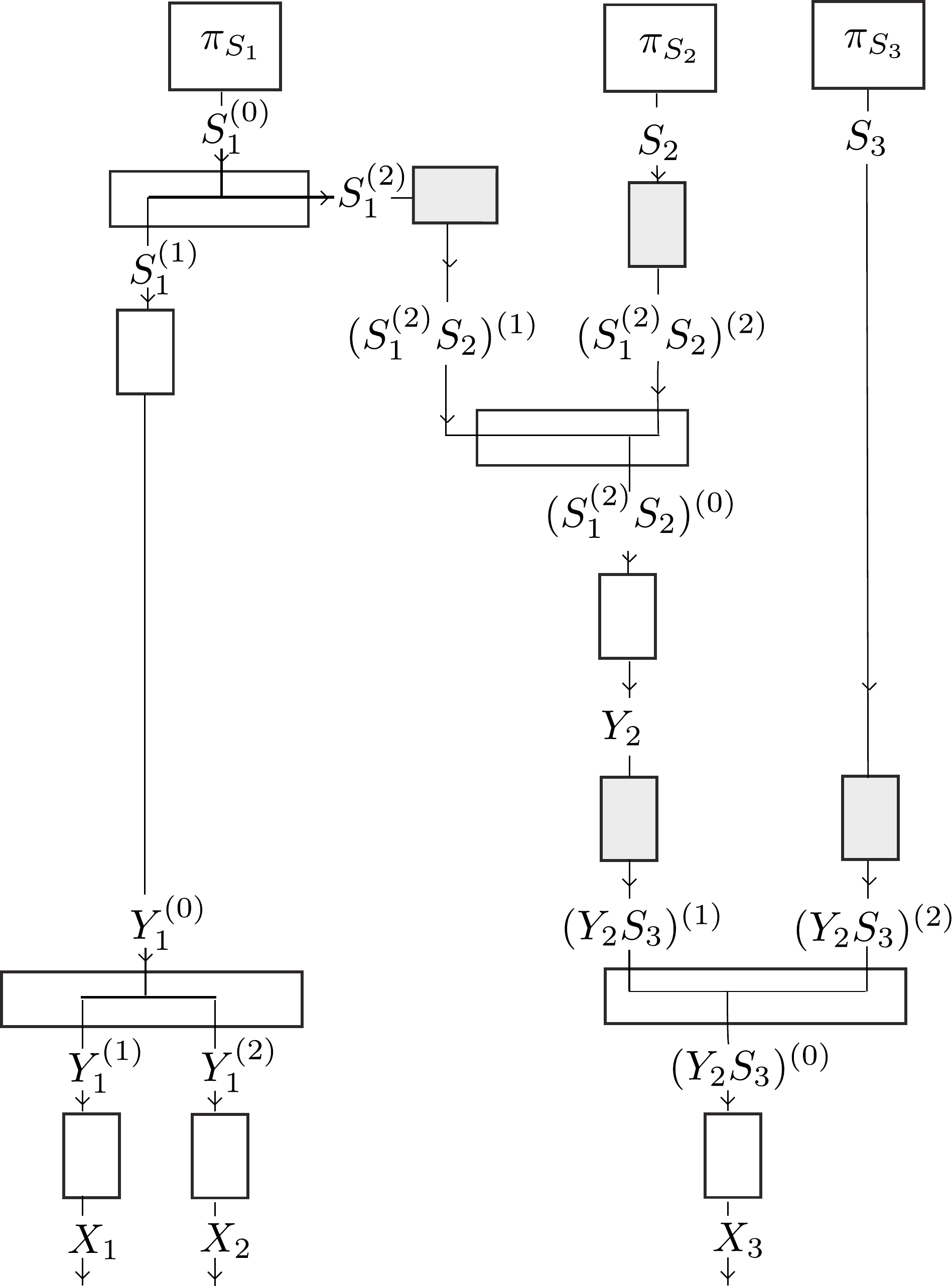}
\caption{Bayesian graph and FGn for a forest with eight variables}
\label{fig:exx4}
\end{figure}

\subsection{Normal Graphs}
\label{sec:ng}

Figures 
\ref{fig:exx1}-\ref{fig:exx4}  show various typical examples of Bayesian graphs and their FGn counterparts. The source blocks are just vectors that represent the priors for the connected variable.  

Note that in an FGn, when a variable feeds more than one child node, it must be replicated as many times as the number of children through a diverter 
(Figure \ref{fig:exx1}). Conversely multiple parents of a child node must be mapped first to a product-space variable, before being fed into the child variable (Figure \ref{fig:exx2}, center). We have shaded some blocks in gray to indicate that they represent fixed matrices (not to be learned). Sometimes, when parent nodes feed a child variable, it is assumed that each parent separately maps to the child variable (Figure \ref{fig:exx2} (right)). In this simplified case only three blocks and a diverter are necessary. Note that such a simplified model may reflect a strong assumption on the separability of the function $P(G|parents(G))$ that cannot account for conditional dependence on configurations of the parent variables.        

To help the reader to fill in the details of the fixed matrices that map to, and from, the product space, we provide some explicit formulas in Appendix A. These expressions are useful also when in the graph of Figure \ref{fig:exx1} the hidden variable represents the whole  product space, in which case the only block to be learned is the prior $\pi_S$. The construction of a tree with embedded product space variables has been presented in \citep{Palmieri_cip2010}.   

It should be noted  that for the variables that are mapped into, or from, their product space,  there may be better representations based on embedding variables with smaller dimensionalities. The product space for multiple variables may be too large to be handled and it is also almost invariably very sparse. The reduction to  smaller latent variables simply requires that the blocks that lead to them have re-defined dimensions and are discovered through learning. 

\subsection{Using the FGn}
\label{sec:use}
\begin{figure}[h]
\center
\includegraphics[width=12.0cm]{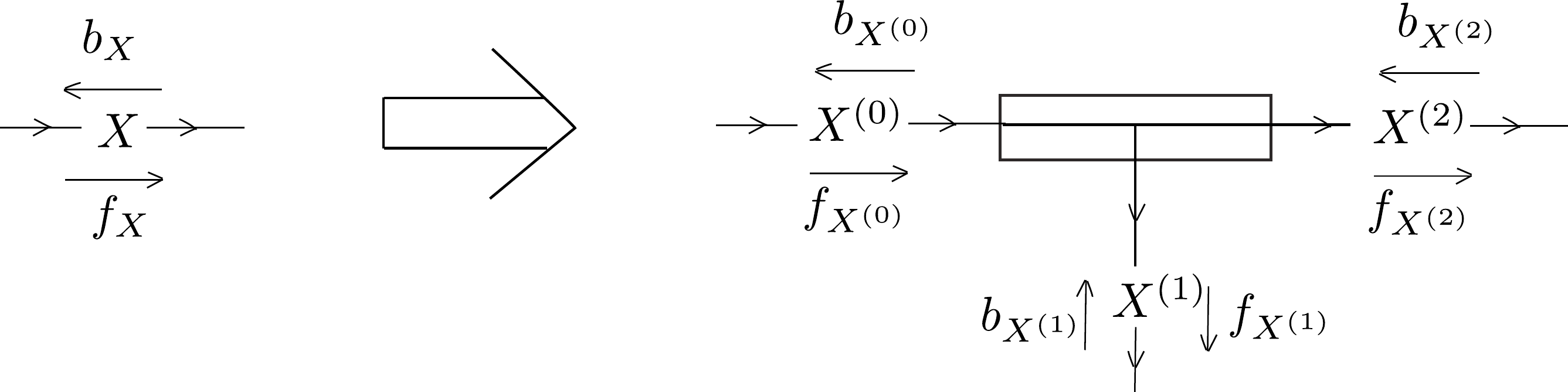}
\caption{Splitting a variable with a diverter block}
\label{fig:split}
\end{figure}

The great feature of Bayesian graphical models is that we can perform exact inference
(at least when there are no loops)  simply by letting the messages propagate and collecting the results. Information can be injected  at the nodes and inference can be computed  for each variable with the product rule. Factor graphs in normal form have the additional advantage of allowing easier modularity with respect to the Bayesian graphs as each variable can be split with a diverter block as shown in Figure \ref{fig:split}.     
The variable to split  can be a hidden,  and/or a source, or a terminal variable. The operation does not alter the rest of the graph and it is like inserting a T-junction in the probability pipeline. Information can be extracted or injected. Extracted information taps from forward and backward flows. Injected knowledge is sent to both directions and can be partial, through smooth distributions, or sharp through delta functions (instantiation). In the latter case the diverter insertion is equivalent to cut the flow (immediate to verify from the product rule). Hence, the diverter block of FGn can be used a fusion bus when new information becomes available from sources that were not included in the original model.  Note that all branches carry always bi-directional information. A given  direction is assigned to each variable simply to define unambiguously  forward and backward flows.

\section{Parametric Learning} 
\label{sec:parl}

\begin{figure}[h]
\center
\includegraphics[width=4.0cm]{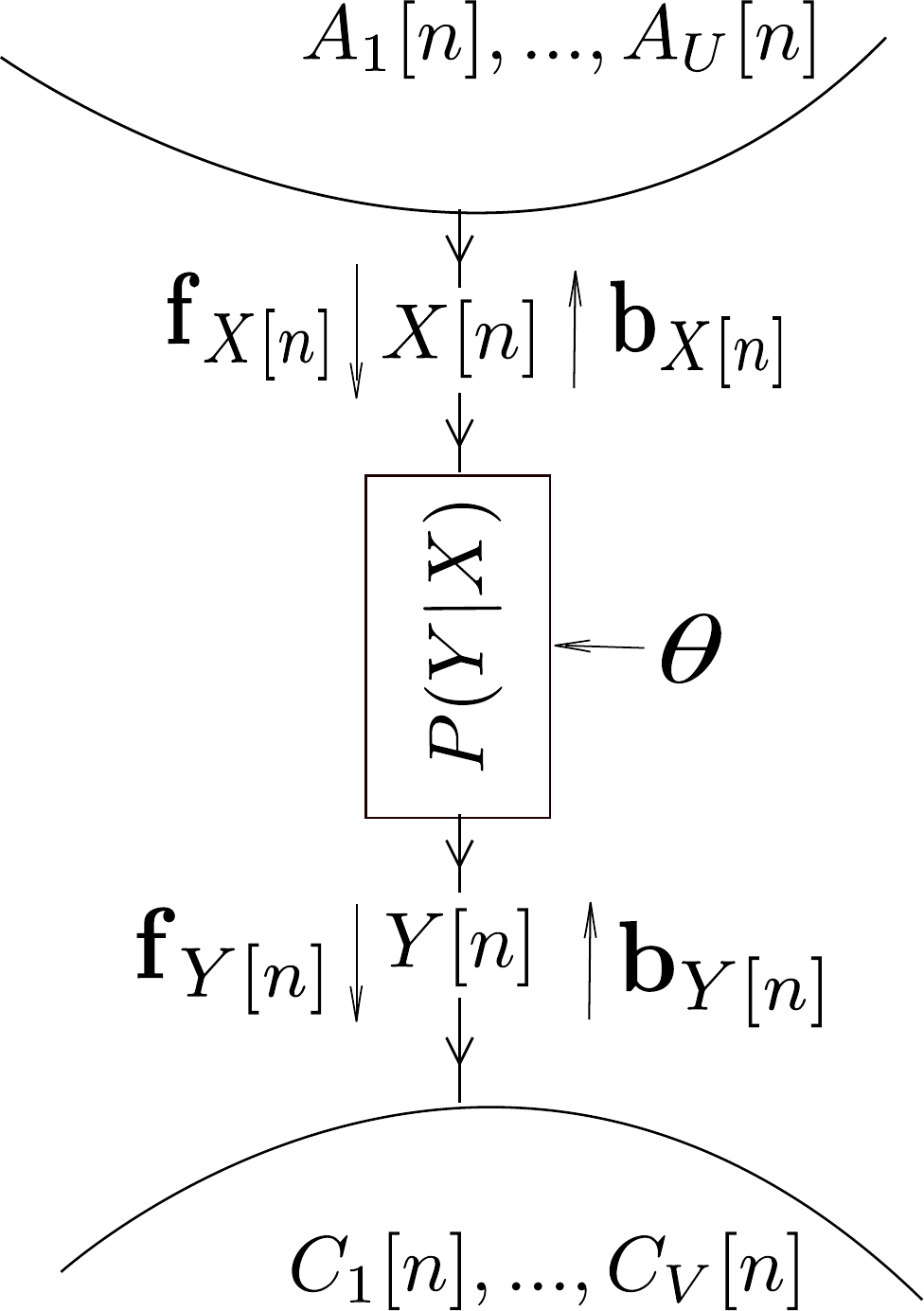}
\caption{A single block embedded in  an FGn}
\label{fig:oneblock_par}
\end{figure}

Figure \ref{fig:oneblock_par} shows a single block  embedded in a cycle-free graph.  The subsystem connects  variable $X$ and $Y$ that  take values in the alphabets ${\cal X}=\{ x^1,x^2,...,x^{M_X}\}$ and  ${\cal Y }=\{ y^1,y^2,...,y^{M_Y}\}$. The block represents the $M_X \times M_Y$ unknown conditional probability matrix $P(Y|X)=[P_r\{Y=y^j|X=x^i \}]=[\theta_{ij}]_{i=1:M_X}^{j=1:M_Y}$. 
There are other variables in the graph  $A_1,...,A_U$, $C_1,...,C_V$ grouped as shown in the figure. Information flows bi-directionally via forward  and backward  messages.
Following the EM algorithm, i.e. assuming that in whole system with fixed parameters all messages have been propagated for a number of steps equal to the graph diameter (E-step), we need to update matrix $\theta$ from independent realization of ${\bf b}_{Y[n]}$ and ${\bf f}_{X[n]}$ for $n=1:N$ (M-step).    

From Figure \ref{fig:oneblock_par}, the total joint pdf can be factorized as 
\begin{equation}
P(XYA_1...A_UC_1...C_V;\theta)=P(C_1...C_V|Y)P(Y|X;\theta)
P(XA_1...A_U). 
\end{equation}
Now if evidence at time $n$  is available  in the graph for $n=1:N$, in the form of instantiated variables somewhere in the system, denoted globally as ${\cal E}[n]$,
the joint likelihood for $X$ and $Y$ at time $n$ can be written as 
\begin{equation}
\begin{array}{l}
p_{X[n]Y[n]{\cal E}[n]}(xy;\theta)=p_{Y|X}(y|x;\theta) \\
~~~~~~\cdot \sum_{a_1...a_U} p_{X[n]A_1[n]...A_U[n]}(xa_1...a_U) 
~\epsilon_{X[n]}(x)~ \epsilon_{A_1[n]}(a_1) \cdot ... \cdot \epsilon_{A_U[n]}(a_U)\\
~~~~~~\cdot \sum_{c_1...c_V} p_{C_1[n]...C_V[n]|Y[n]}(c_1...c_V|y)
~\epsilon_{C_1[n]}(c_1)\cdot ... \cdot \epsilon_{C_V[n]}(c_v) ~\epsilon_{Y[n]}(y) ,
\end{array}
\end{equation}   
where the {\em evidence} function $\epsilon$ for a variable $Z[n]$ is defined as 
\begin{equation}
\epsilon_{Z[n]}(z)= \left\{ 
\begin{array}{l}
\delta(z-\overline{z}[n])~~{\rm if} ~ Z ~{\rm at~ time}~ n ~{\rm is~ fixed~ to ~ value} ~\overline{z}[n], \\
1 ~~~{\rm otherwise} . 
\end{array}          \right.
\end{equation} 
Note that in the model the set of instantiated variables could be different for each $n$. Also $X$ and/or $Y$ are among the variables that could be fixed to specific values.  Therefore the block could be  a  hidden, a terminal, or a source factor. When it is a source block, either erase completely the variables $XA_1...A_U$ from the equations, or assume that the input flow is a constant uniform distribution.    

Assuming independent samples, the empirical likelihood becomes 
\begin{equation}
L(\theta)=\prod_{n=1}^N \sum_{x} \sum_{y} p_{X[n]Y[n]{\cal E}[n]}(xy;\theta)
=\prod_{n=1}^N \sum_{x} \sum_y f_{X[n]}'(x) p_{Y|X}(y|x;\theta)b_{Y[n]}'(y), 
\label{eq:lik}
\end{equation}
where 
\begin{equation}
\begin{array}{l}
f_{X[n]}'(x)=\sum_{a_1...a_U} p_{X[n]A_1[n]...A_U[n]}(xa_1...a_U) 
~\epsilon_{X[n]}(x)~ \epsilon_{A_1[n]}(a_1) \cdot ... \cdot \epsilon_{A_U[n]}(a_U) ,\\
b_{Y[n]}'(y) =\sum_{c_1...c_V} p_{C_1[n]...C_V[n]|Y[n]}(c_1...c_V|y)
~\epsilon_{C_1[n]}(c_1)\cdot ... \cdot \epsilon_{C_V[n]}(c_v) ~\epsilon_{Y[n]}(y) ,
\end{array}
\end{equation}
are functions proportional to the forward and backward messages reaching the block. 
Note that neither 
$f_{X[n]}'(x)$ nor $b_{Y[n]}'(y)$  are necessarily distributions (sum to one). In handling Bayesian graphical models  is generally up to the designer to keep the propagating messages normalized 
(even though it is a good practice for numerical stability). Therefore what reaches our block are in general messages that are proportional to $f_{X[n]}'(x)$ and $b_{Y[n]}'(y)$
\begin{equation}
f_{X[n]}(x) = K_{f_{X{[n]}}} f_{X[n]}'(x);~~~~ b_{Y[n]}(y) = K_{b_{Y{[n]}}} b_{Y[n]}'(y),
\end{equation}
where $ K_{f_{X{[n]}}}$ and $K_{b_{Y{[n]}}}$ are constants that may be different at each time step. In the following we will assume that both $f_{X[n]}(x)$ and  $b_{Y[n]}(y)$ are kept normalized to be valid distributions. 
Equation (\ref{eq:lik}) for the likelihood can be re-written using the vector-matrix notation as  
\begin{equation}
L(\theta) =\prod_{n=1}^N K_{f_{X{[n]}}} {\bf f}_{X[n]}^T~ \theta~ {\bf b}_{Y[n]}K_{b_{Y{[n]}}}.
\label{eq:likvm}
\end{equation}
The log-likelihood is  then
\begin{equation}
\ell(\theta) =\log(L(\theta))=\sum_{n=1}^N   \log \left( {\bf f}_{X[n]}^T~ \theta~ {\bf b}_{Y[n]} \right) + \sum_{n=1}^N \log \left(K_{f_{X{[n]}}} K_{b_{Y{[n]}}} \right),
\label{eq:logl}
\end{equation}
where only the first term depends on $\theta$. 

Various algorithms for updating matrix $\theta$ have been derived in the literature. Some look very similar to each other (see for example Appendix B in \citep{Harmeling2011}, or Baum-Welch algorithm for HMMs \citep{Welch2003}). Some are derived from good heuristics and some from different cost functions. We found that apparently minor different normalizations, that are often implementation-dependent, play a crucial role in the quality of the solutions in terms of likelihood when forward and backward messages are not delta functions. In the following we  present three different approaches for deriving  parametric learning algorithms. 
A Bayesian model for learning the block parameters is presented in Section \ref{sec:bay}.  

\subsection{Maximum Likelihood Learning}
To derive an algorithm that maximizes the log-likelihood (\ref{eq:logl}),  the problem can be directly cast into the following constrained optimization  scheme 
\begin{equation}
\left\{ \begin{array}{l}
\min_{\theta} ~~ - \sum_{n=1}^N L[n] \log \left( {\bf f}_{X[n]}^T~ \theta~ {\bf b}_{Y[n]} \right), \\
\theta ~~~~ {\rm row-stochastic},
\end{array} \right. 
\label{eq:mlp}
\end{equation}
where we have included a learning mask $L[n]$ that  is set to one on the training set and zero elsewhere. In Appendix C,  we add a stabilizing term to the cost function and apply KKT conditions to obtain the following algorithm.
\bigskip
\hrule
\bigskip
\noindent
{\bf ML Algorithm: }
\smallskip

\noindent
(0) Initialize $\theta$ to uniform rows: $\theta=(1/M_Y) {\bf 1}_{M_X \times M_Y}$

\noindent
(1) $\theta_{lm} \longleftarrow {\theta_{lm} \over \sum_{n=1}^N L[n] f_{X[n]}(l)} \sum_{n=1}^N L[n] {  f_{X[n]}(l) b_{Y[n]}(m) \over {\bf f}_{X[n]}^T \theta {\bf b}_{Y[n]} }$,~~~  $l=1:M_X$, $m=1:M_Y$.

\noindent
(2) Row-normalize $\theta$ and go back to (1)

\bigskip
\hrule 
\bigskip
\noindent
{\footnotesize 
(We have used the shortened notation $f_{X[n]}(x^l)=f_{X[n]}(l)$,  $b_{Y[n]}(y^m)=b_{Y[n]}(m)$).}  
\bigskip

\noindent
The algorithm is a fast multiplicative update with no free parameters. The iterations usually converge in a few steps and the presence of the normalizing factors  makes the algorithm numerically very stable.  The conditions on local stability discussed in Appendix C, guarantee uniform convergence to a local minimum.

The ML algorithm can also be derived observing that 
\begin{equation}
\theta_{ml}= {  P_{YX}(y^mx^l) \over P_X(x^l) }, ~~~l=1:M_X,~m=1:M_Y.
\end{equation}
The numerator and the denominator can be estimated from samples as
\begin{equation}
P_{YX}(y^mx^l) \simeq {1 \over N} \sum_{n=1}^N 
{ p_{YX {\cal E}[n]}(y^m x^l; \theta )   \over 
 \sum_{i=1}^{M_X} \sum_{j=1}^{M_Y}p_{YX {\cal E}[n]}(y^j x^i; \theta )}=
{1 \over N} \sum_{n=1}^N { b_{Y[n]}(m)  \theta_{lm} f_{X[n]}(l)  \over 
 \sum_{i=1}^{M_X} \sum_{j=1}^{M_Y} b_{Y[n]}(j)  \theta_{ij} f_{X[n]}(i)};
\end{equation}
\begin{equation}
P_X(x^l) \simeq {1 \over N} \sum_{n=1}^N f_{X[n]}(l).
\end{equation}
Now $\theta_{lm}$ appears on both sides of the equation
\begin{equation}
\theta_{lm} \simeq {1 \over \sum_{n=1}^N f_{X[n]}(l)} \sum_{n=1}^N { b_{Y[n]}(m)  \theta_{lm} f_{X[n]}(l)  \over 
 \sum_{i=1}^{M_X} \sum_{j=1}^{M_Y} b_{Y[n]}(j)  \theta_{ij} f_{X[n]}(i)},
\end{equation}
that immediately gives the ML recursions. The recursion is very similar to Baum-Welch algorithm,  used to train
Hidden Markov Models (HMM)   \citep{Welch2003}. The  algorithm  was initially derived heuristically and then proven to provide a monotonic increase in the likelihood for the HMM model. The derivation is quite typical in many papers on graphical models where the recursions are derived from writing down the joint probability function for the variables at the ends of each branch (see for example  \citep{Harmeling2011}).

\subsection{Minimum Divergence Learning}

Another algorithm, similar to ML,  can be derived by recognizing that  the likelihood (\ref{eq:likvm}) can be written as 
\begin{equation}
L(\theta)=\prod_{n=1}^N    \left( {\bf f}_{Y[n]}^T~ {\bf b}_{Y[n]} \right) K[n],
\end{equation}
where ${\bf f}_{Y[n]}=\theta^T {\bf f}_{X[n]}$ is the forward distribution for $Y[n]$, and $K[n]$ is a constant that does not depend on $\theta$.  It seems natural to attempt to match ${\bf f}_{Y[n]}^T$ as much as possible to $ {\bf b}_{Y[n]}$ to maximize $L(\theta)$.  We can formulate the problem by seeking the minimization of the average KL-divergence 
$KL({\bf b}_{Y[n]} || {\bf f}_{Y[n]})$ over the samples, i.e. minimize
\begin{equation}
d(\theta)=\sum_{n=1}^N L[n] \sum_{i=1}^{M_y} b_{Y[n]}(j) \log { b_{Y[n]}(j)  \over \sum_{i=1}^{M_X} \theta_{ij} f_X[n](i)  }.
\label{eq:div}
\end{equation}
This cost function has been used for approximation of  non linear functions through density matching \citep{PalmieriTSP2013}. The idea is that the backward message $b_{Y[n]}$ acts as the {\em desired} density and the block has to provide its best estimate $f_{Y[n]}$. The criterion does not exactly maximize the likelihood, but only its lower bound. To see this,  since $b_{Y[n]}(y)$ is a distribution, we can use Jensen's inequality to get
\begin{equation}
\begin{array}{l}
\sum_{n=1}^N \log \sum_{y} b_{Y[n]}(y) f_{Y[n]}(y) \ge 
\sum_{n=1}^N \sum_y   b_{Y[n]}(y) \log f_{Y[n]}(y) \\
= -   \sum_{n=1}^N \left( KL\left( b_{Y[n]}|| f_{Y[n]} \right) -\sum_y b_{Y[n]}(y) \log b_{Y[n]}(y) \right).
\end{array}
\end{equation} 
Therefore the minimization of $d(\theta)$ is equivalent to the maximization of a lower bound of $\ell(\theta)$. Other similar bounds can be obtained by exchanging the role of
$b_{Y[n]}$ and $f_{Y[n]}$, or  by writing out  $f_{Y[n]}$. They are omitted for brevity also because they do not seem to lead to useful algorithms. 

Minimization of (\ref{eq:div}) can be cast again as a constrained optimization problem  and solved using KKT conditions. The derivations follow  steps similar to the ones used for the ML algorithm and are reported in the appendix  of \citep{PalmieriTSP2013}. The
algorithm is derived minimizing a  generalized version of the  KL-divergence  
\begin{equation}
\sum_{n=1}^N L[n] \left[ \sum_{j=1}^{M_Y} b_{Y[n]}(j) \log { b_{Y[n]}(j)  \over \sum_{i=1}^{M_X} \theta_{ij} f_{X[n]}(i)  } + \sum_{j=1}^{M_Y} \sum_{i=1}^{M_X} \theta_{ij} f_{X[n]}(i)   \right],
\end{equation}
where the extra term acts as a stabilizer during convergence and does not affect the final value of the cost function since it becomes one when $\theta$ is normalized to be row-stochastic. 
\bigskip
\hrule
\bigskip
\noindent
{\bf KL Algorithm: }
\smallskip

\noindent
(0) Initialize $\theta$ to uniform rows: $\theta=(1/M_Y) {\bf 1}_{M_X \times M_Y}$

\noindent
(1) $\theta_{lm} \longleftarrow {\theta_{lm} \over \sum_{n=1}^N L[n] f_{X[n]}(l)} \sum_{n=1}^N L[n] {  f_{X[n]}(l) b_{Y[n]}(m) \over \sum_{i=1}^{M_X} \theta_{im} f_{X[n]}(i)}$, ~~~ $l=1:M_X$, $m=1:M_Y$.

\noindent
(2) Row-normalize $\theta$ and back to (1).

\bigskip
\hrule 
\bigskip
This algorithm can be seen as a special case of a well-known iteration to find  Non Negative Matrix Factorization (NNMF) \citep{Ngoc-DiepHo2008} \citep{DanielD.Lee2001}. The updates are similar to the ML algorithm, but use a different normalization.  It is interesting to see that the normalization in the ML algorithm means that each estimate $\theta_{lm}$, during adaptation, competes with the whole matrix, while in the KL algorithm each $\theta_{lm}$ competes only column-wise.   

\subsection{Viterbi Approximation}
Observing that backward and forward messages carry soft score information,  a Viterbi-type of approximation could be obtained by sharpening the backward and forward distributions \citep{Ghahramani2012}  and computing the average outer products (expected counts). Another expected-count approximation is the VAR algorithm that will be presented in Section \ref{sec:bay}.   
Defining $I_{Max}({\bf p})$ to be  the vector with all zeros except the one element set to one that corresponds to the maximum, 
the  algorithm is as  follows.  
\bigskip
\hrule
\bigskip
\noindent
{\bf VIT Algorithm: }
\smallskip

\noindent
(0) Define a very small positive value $\delta$;

\noindent
(1) ${\bf e}_{X[n]}=I_{Max}({\bf f_{X[n]}}) + \delta {\bf 1}_{M_X \times 1}$; ${\bf e}_{Y[n]}= I_{Max}({\bf b_{Y[n]}}) + \delta {\bf 1}_{M_Y \times 1}$;

\noindent
(2) $\theta=  \sum_{n=1}^N L[n] {\bf e}_{X[n]} {\bf e}_{Y[n]}^T$;

\noindent 
(3) Row-normalize $\theta$. 
\bigskip
\hrule 
\bigskip
\noindent
In the algorithm we have added a small positive constant to all the elements  to avoid the null values that may cause divisions by zero in the normalization. Viterbi approximation essentially assumes that occurrence counting can be done on the maximum likelihood estimates. This approach  is expected to perform comparably to the other algorithms  when forward and backward distributions are sharp, or when we have available a very large number of samples.   

\section{Bayesian Learning}
\label{sec:bay}
\begin{figure}[ht]
\center
\includegraphics[width=15.0cm]{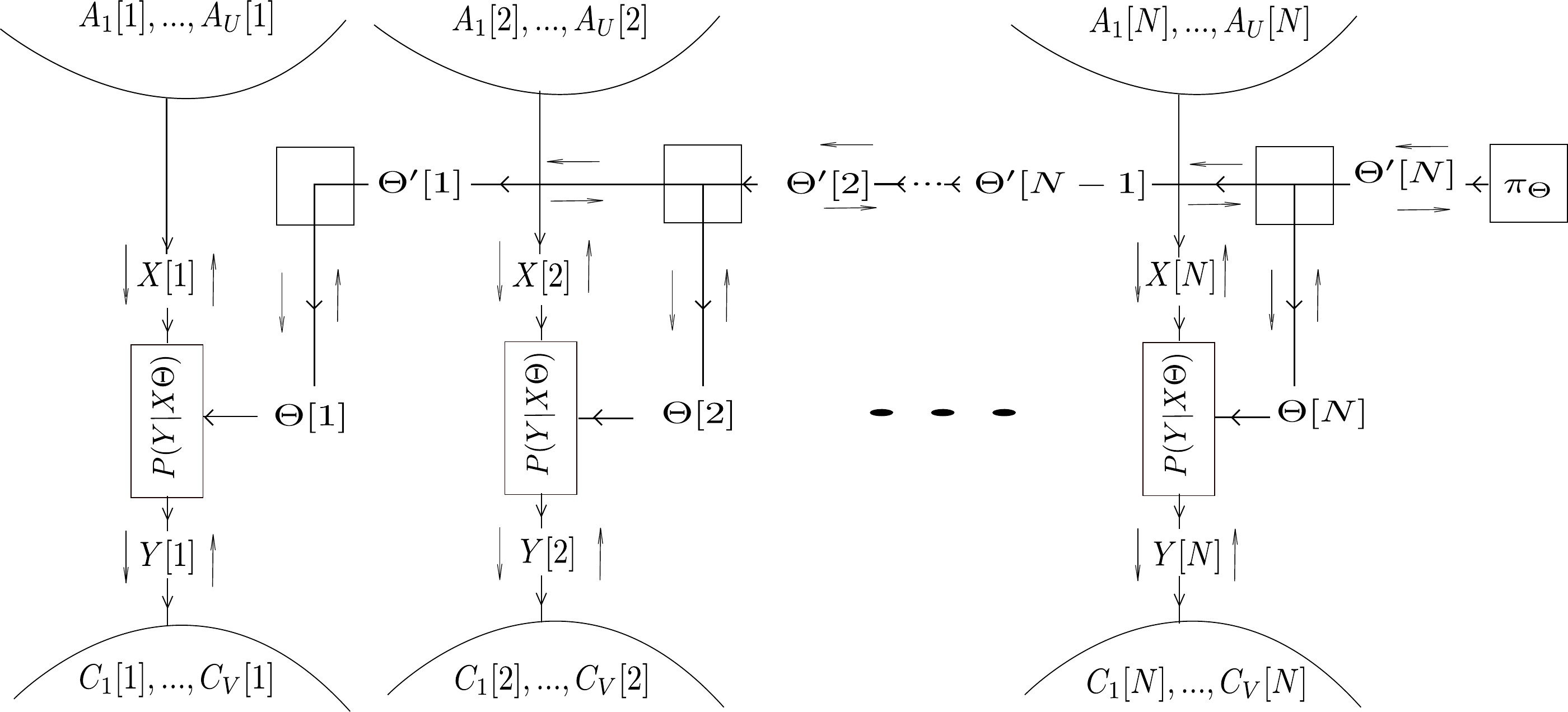}
\caption{The complete factor graph for the Bayesian approach.}
\label{fig:oneblock_bay}
\end{figure}

In formulating the learning problem in a Bayesian framework, we  assume that $\Theta$ is a random matrix that belongs to the space ${\cal T}$ of $M_X \times M_Y$  row-stochastic matrices.  The $N$ independent samples, collected anywhere in the system and translated in traveling messages, must be optimally used to provide an estimate for $\Theta$. We lay down the  problem in the factor graph of  
Figure \ref{fig:oneblock_bay}, where we show $N$ independent realizations of the model that share the  variable $\Theta$.
The variable $\Theta[n]$ is related to the $n$-th realization that can exchange  information  with the other realizations. There is also a block $\pi_\Theta$ that represents  prior knowledge about $\Theta$. The information flow is carried by forward and backward messages 
$f_{\Theta[n]}(\theta)$ and $b_{\Theta[n]}(\theta)$, which are  matrix functions. Following the standard rules of belief propagation, the messages are  related to each other via marginalization as 
\begin{equation}
\begin{array}{l}
f_{Y[n]}(y) \propto \int_{\theta \in {\cal T}}\sum_{x \in {\cal X}} P_{Y|X\Theta}(y|x \theta) f_{X[n]}(x) f_{\Theta[n]}(\theta) d\theta; \\ 
b_{X[n]}(x) \propto \int_{\theta \in {\cal T}}\sum_{y \in {\cal Y}} P_{Y|X\Theta}(y|x \theta) b_{Y[n]}(y) f_{\Theta[n]}(\theta) d \theta; \\
b_{\Theta[n]}(\theta) \propto \sum_{x \in {\cal X}}\sum_{y \in {\cal Y}} P_{Y|X\Theta}(y|x \theta) b_{Y[n]}(y) f_{X[n]}(x).
\end{array}
\end{equation}
In vector-matrix notation we can write 
\begin{equation}
\begin{array}{l}
{\bf f}_{Y[n]} \propto \int_{\theta \in {\cal T}} \theta^T {\bf f}_{X[n]} f_{\Theta[n]}(\theta) d \theta = F_{\theta[n]}^T {\bf f}_{X[n]}; \\
{\bf b}_{X[n]} \propto \int_{\theta \in {\cal T}} \theta {\bf b}_{Y[n]} f_{\Theta[n]}(\theta) d \theta = F_{\theta[n]} {\bf b}_{Y[n]};
\end{array}
\end{equation}
where $F_{\theta[n]}=\int_{\theta \in {\cal T}} \theta f_{\Theta[n]}(\theta) d \theta$ is the mean forward matrix for $\Theta[n]$. The backward message for $\Theta[n]$ is the matrix function 
\begin{equation}
b_{\Theta[n]}(\theta) \propto {\bf f}_{X[n]}^T \theta {\bf b}_{Y[n]}= {\bf f}_{X[n]}^T 
\left( \begin{array}{ccc}
\theta_{11} & ... & \theta_{1M_Y} \\
\theta_{21} & ... & \theta_{2M_Y} \\
. & ... & . \\
\theta_{M_X1} & ... & \theta_{M_XM_Y} 
\end{array} \right) {\bf b}_{Y[n]}.
\end{equation}
Messages for $\Theta[n]$ and $\Theta'[n]$ in the other branches are  the result of the product rule 
$f_{\Theta[n]}(\theta) \propto f_{\Theta'[n]}(\theta) b_{\Theta'[n-1]}(\theta)$;
$b_{\Theta'[n]}(\theta) \propto b_{\Theta[n]}(\theta) b_{\Theta'[n-1]}(\theta)$;
$f_{\Theta'[n]}(\theta) \propto b_{\Theta[n+1]}(\theta) f_{\Theta'[n+1]}(\theta)$. Each message about $\Theta$ is a product of the type 
\begin{equation}
\mu_{\Theta}(\theta) \propto \prod_l {\bf f}_{X[l]}^T \left( \begin{array}{ccc}
\theta_{11} & ... & \theta_{1M_Y} \\
\theta_{21} & ... & \theta_{2M_Y} \\
. & ... & . \\
\theta_{M_X1} & ... & \theta_{M_XM_Y} 
\end{array} \right) {\bf b}_{Y[l]} =
\prod_l \sum_{i=1}^{M_X} \sum_{j=1}^{M_Y} b_{Y[l]}(y^j) f_{X[l]}(x^i) \theta_{ij}
\label{eq:intr}
\end{equation}
If variables $X[n]$ and $Y[n]$ of block $n$ are both instantiated, i.e. forward and backward messages are delta functions,  $f_{X[n]}(x)=\delta(x-x^i)$, $b_{Y[n]}(x)=\delta(y-y^j)$, the backward message  from block $n$ is simply $b_{\Theta[n]}(\theta) \propto \theta_{ij}$. If also all variables from all $n$ are instantiated, information exchanged among the blocks (except possibly for the prior on $\Theta$) are exactly products of Dirichlet distributions    
\begin{equation}
\mu_{\Theta}(\theta) \propto \prod_{i=1}^{M_X} \prod_{j=1}^{M_Y} \theta_{ij}^{n_{ij}} \propto 
\prod_{i=1}^{M_X} Dir(\theta_{i1},...,\theta_{iM_Y};n_{i1}+1,...,n_{iM_Y}+1),
\end{equation}
where $n_{ij}$ are the integer numbers that represent the cumulative counts of the occurrences of pair $(i,j)$ (hard scores). 
Unfortunately, in the general case in which forward and backward messages carry soft information, expression (\ref{eq:intr}) becomes intractable.
This is because any estimate of $\theta$ would require handling the product of multiple sums of all the elements of $\theta$. For example: the mean of an expression like  (\ref{eq:intr}) would require multiple integration on $\cal T$; the computation of the argmax over $\cal T$ (MAP estimate) would require a search on a very large parameter space. Therefore, the Bayesian formulation, to become tractable,  requires that we resort to some kind of approximation. 

\subsection{Variational Approximation}

Variational methods are very popular in graphs for approximating intractable densities \citep{Dauwels2007}\citep{Beal_Gha2004}\citep{Winn_Bish2005}. If we focus on the message for $\Theta[n]$ that leaves the block $P(Y|X\Theta)$, applying the independent  variational approximation \citep{Dauwels2007}, we get that the estimate $b_{\Theta[n]}^V(\theta)$ of $b_{\Theta[n]}(\theta)$ is 
\begin{equation}
\begin{array}{l}
b_{\Theta[n]}^V(\theta) \propto e^{\sum_{i=1}^{M_X} \sum_{j=1}^{M_Y} b_{Y[n]}(y^j) f_{X[n]}(x^i) \log \theta_{ij}} = \prod_{i=1}^{M_X} \prod_{j=1}^{M_Y} \theta_{ij}^{b_{Y[n]}(y^j) f_{X[n]}(x^i)} \\ 
\propto 
\prod_{i=1}^{M_X} Dir(\theta_{i1},...,\theta_{iM_Y};f_{X[n]}(x^i) b_{Y[n]}(y^1) +1,..., f_{X[n]}(x^i) b_{Y[n]}(y^{M_Y}) +1), 
\end{array}
\end{equation}
which is again the product of $M_X$ Dirichlet distributions. The Dirichlet distribution, sometimes used as an assumption 
\citep{Heckerman1996}\citep{Winn_phd2004}, is exactly the variational approximation. Taking also the distribution $\pi_\Theta(\theta)$ to be a product of Dirichlet densities
\begin{equation}
\pi_{\Theta}(\theta) \propto 
\prod_{i=1}^{M_X} Dir(\theta_{i1},...,\theta_{iM_Y};\alpha_{i1}+1,...,\alpha_{iM_Y}+1),
\end{equation}
with $\alpha_{ij}$  our a priori knowledge about $\theta_{ij}$, all messages in the upper branches have the form 
\begin{equation}
\begin{array}{l}
\mu_{\Theta}(\theta) \propto 
\prod_{i=1}^{M_X} Dir \left( \theta_{i1},...,\theta_{iM_Y}; \right.\\
~~~ \left. \alpha_{i1}+\sum_l f_{X[l]}(x^i) b_{Y[l]}(y^1) +1,..., \alpha_{iM_Y}+\sum_l f_{X[l]}(x^i) b_{Y[l]}(y^{M_Y}) +1
\right). 
\end{array}
\end{equation}
This approximation essentially implies that all the entries of  $\Theta[n]$ are
independent variables constrained to be in $[0, 1]$ and that row-sum to one.    

Comprehensive knowledge  about $\Theta$ can be collected  at the end of the $N$-th stage using the product rule 
\begin{equation}
p_{\Theta'[N]}(\theta) \propto f_{\Theta'[N]}(\theta) b_{\Theta'[N]}(\theta). 
\end{equation}    
This is clearly a product of  Dirichlet distributions for which we can obtain our estimate for $\theta$ computing either the mean or the mode.
Choosing to take the mode (the mean is very similar) we get that our best estimate for the parameters is 
\begin{equation}
\theta_{ij} \propto \alpha_{ij}+\sum_{n=1}^N f_{X[n]}(x^i) b_{Y[n]}(y^j).
\end{equation}    
The above  expression can be immediately translated in the the following algorithm.   
\bigskip
\hrule
\bigskip
\noindent
{\bf VAR Algorithm: }

\smallskip
\noindent
(0) Define a small positive value $\delta$;

\noindent
(1) $\theta_{lm} \longleftarrow \delta + \sum_{n=1}^N L[n]{  f_{X[n]}(l) b_{Y[n]}(m) }$,  $l=1:M_X$, $m=1:M_Y$;

\noindent
(2)  Row-normalize $\theta$.

\bigskip
\hrule 
\bigskip
In the algorithm we have assumed that we have no previous knowledge about $\theta$
setting the initial matrix to uniform rows. The inclusion of the small constant $\delta$ is also useful to avoid divisions by zero in the normalizations if certain matrix rows are never filled by the outer products.  

The variational approximation reflects a very common  heuristic: take $\theta$ to be just the average on the training set of the outer products of backward and forward messages. 
The estimate can be seen as the normalized sum of {\em soft scores}, or {\em expected scores}. Many proposals in learning graphical models end up in averaging soft scores or score estimates  \citep{Ghahramani2012} \citep{Koller2010}.   

The time-unfolded graph of Figure \ref{fig:oneblock_bay} has been used in \citep{Palmieri2012a} for an application to a typical control example with the VAR algorithm. When we are at time $n$ and only partial information is  available, the graph can provide a simple framework 
for prediction  and/or  smoothing, including in a unique framework training and testing. We leave such a discussion to another paper having limited ourselves here to the issue of   the best estimate for $\Theta$ from $N$ examples.   

\section{General Considerations}

Algorithms ML and KL are iterative in nature, while the other two, VIT and VAR,  require only one step (batch algorithms). It is up to the designer to decide how many iterations to use for each block for ML and KL within the EM framework. Usually this is achieved by trial and error, even if we know that the more iterations we use, the more each block will act in a greedy maner in maximizing its localized  likelihood estimation. In the following we will provide some experimental result that should be of some guidance to the programmer. 

We would like to emphasize again that all the four algorithms, if backward and forward messages are delta functions, are really the same algorithm reduced to simple  co-occurrence counting.

\section{Simulations on a Single Block}
\label{sec:single}
In this first set of experiments we have simulated learning in a single block on randomly generated forward and backward distributions ${\bf f}_{X[n]}$ and ${\bf b}_{Y[n]}$, $n=1:N$. The idea is that if learning is embedded in an arbitrary graph, it should cope with all sorts of distributions. The distributions we have used are all independent samples from a uniform number generator in $[0,1]$ normalized to sum one for each $n$: ${\bf u}_{X[n]}$, ${\bf u}_{Y[n]}$. To generate sharper distributions we have raised each distribution value to a large exponent,   
${\bf f}_{X[n]} \propto {\bf u}_{X[n]}$$.\wedge EX$, ${\bf b}_{Y[n]} \propto {\bf u}_{Y[n]}$$.\wedge EY$, and normalized them. When the exponent is very large, the distributions become very similar to delta distributions. 

\begin{figure}[ht]
\center
\includegraphics[width=7.0cm]{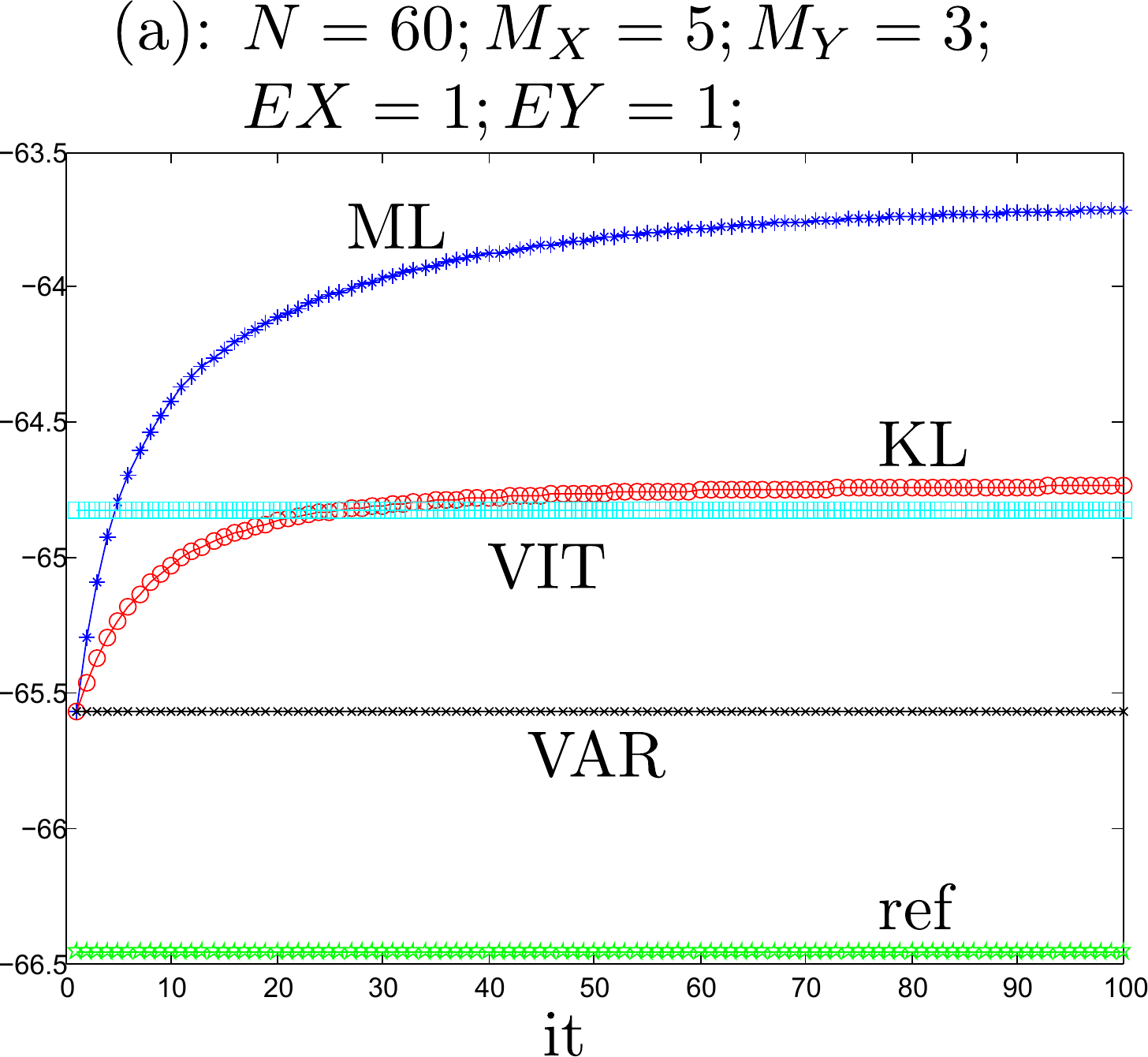}
\includegraphics[width=7.0cm]{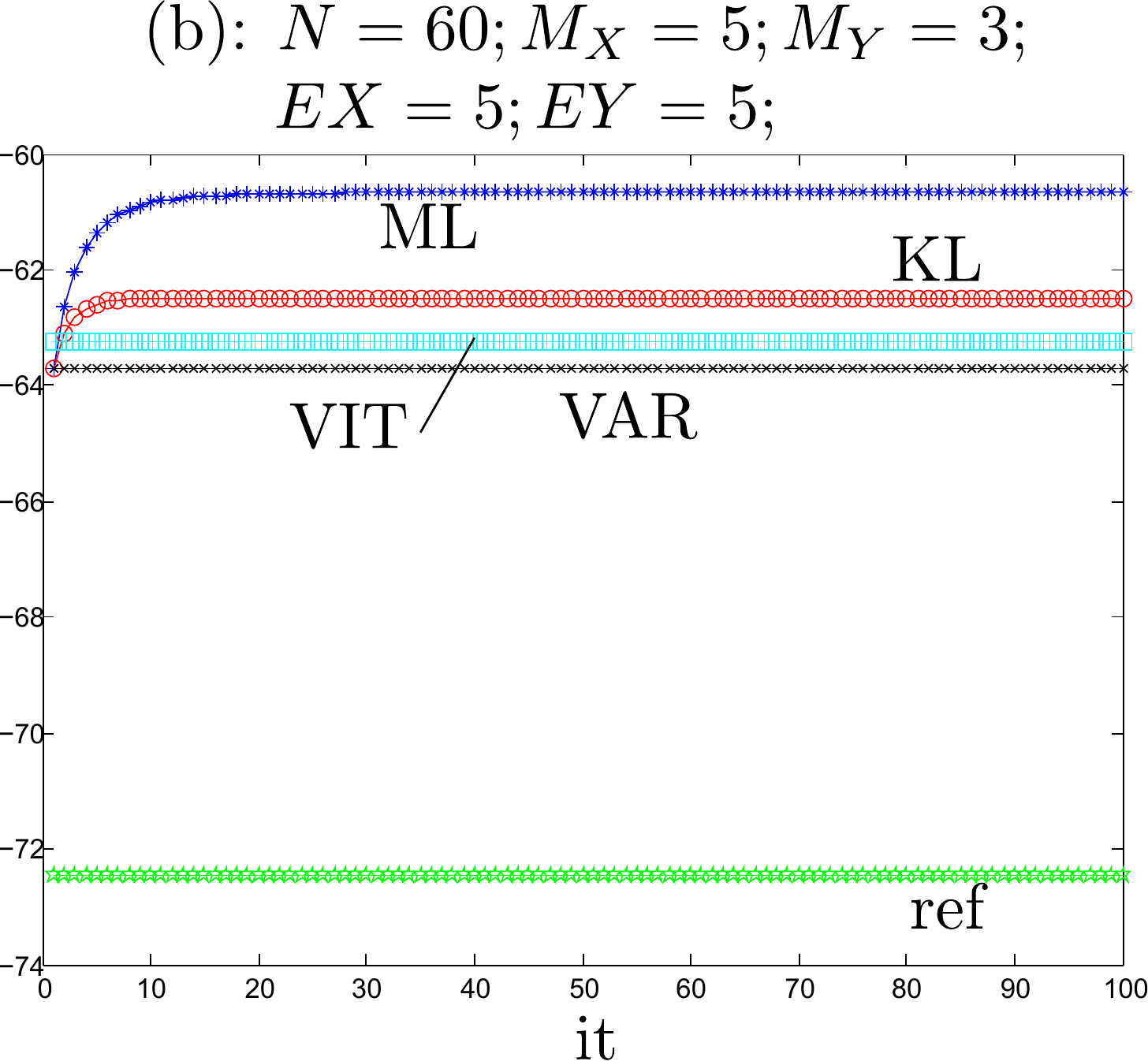} \\
\bigskip
\includegraphics[width=7.0cm]{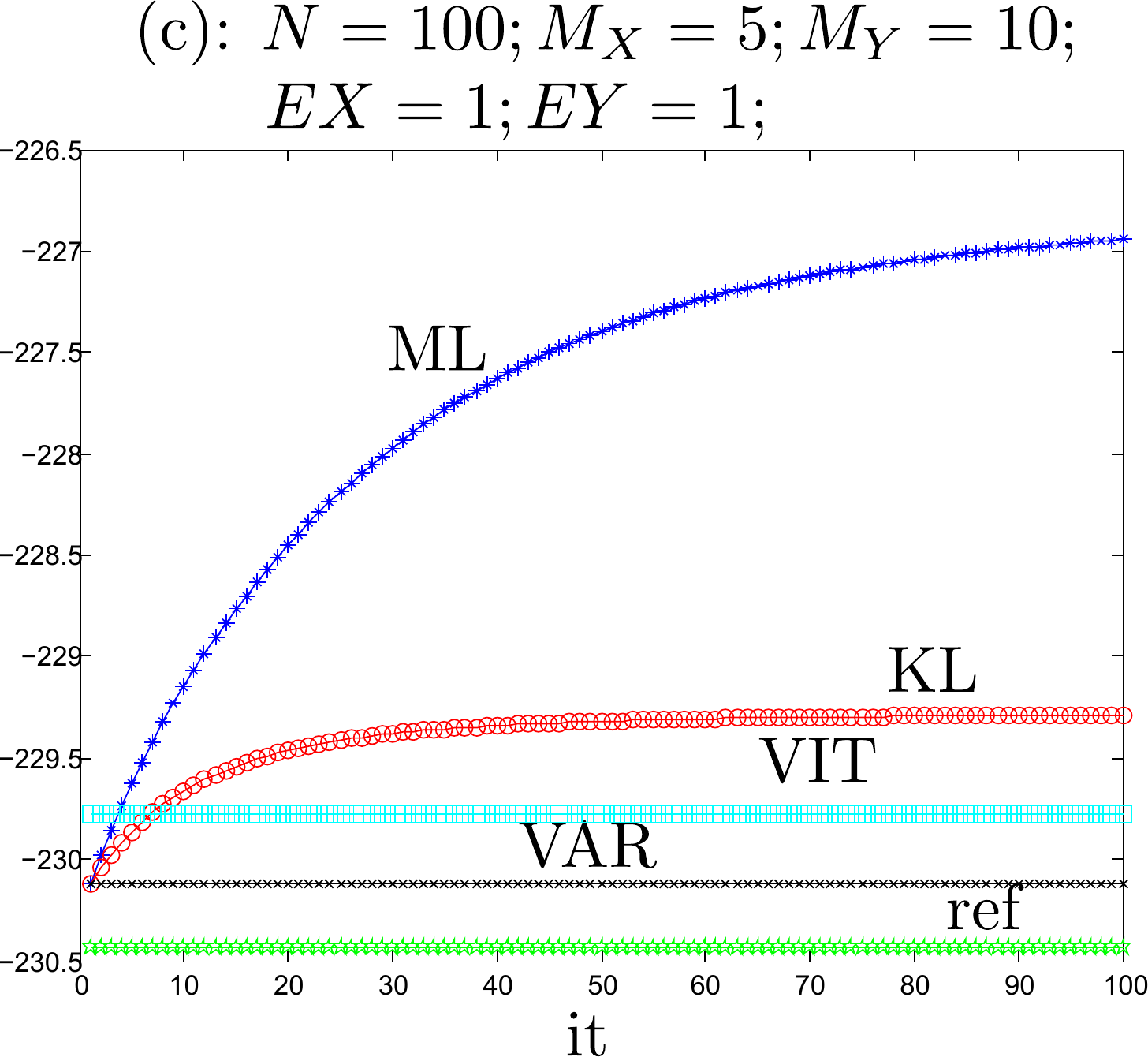}
\includegraphics[width=7.0cm]{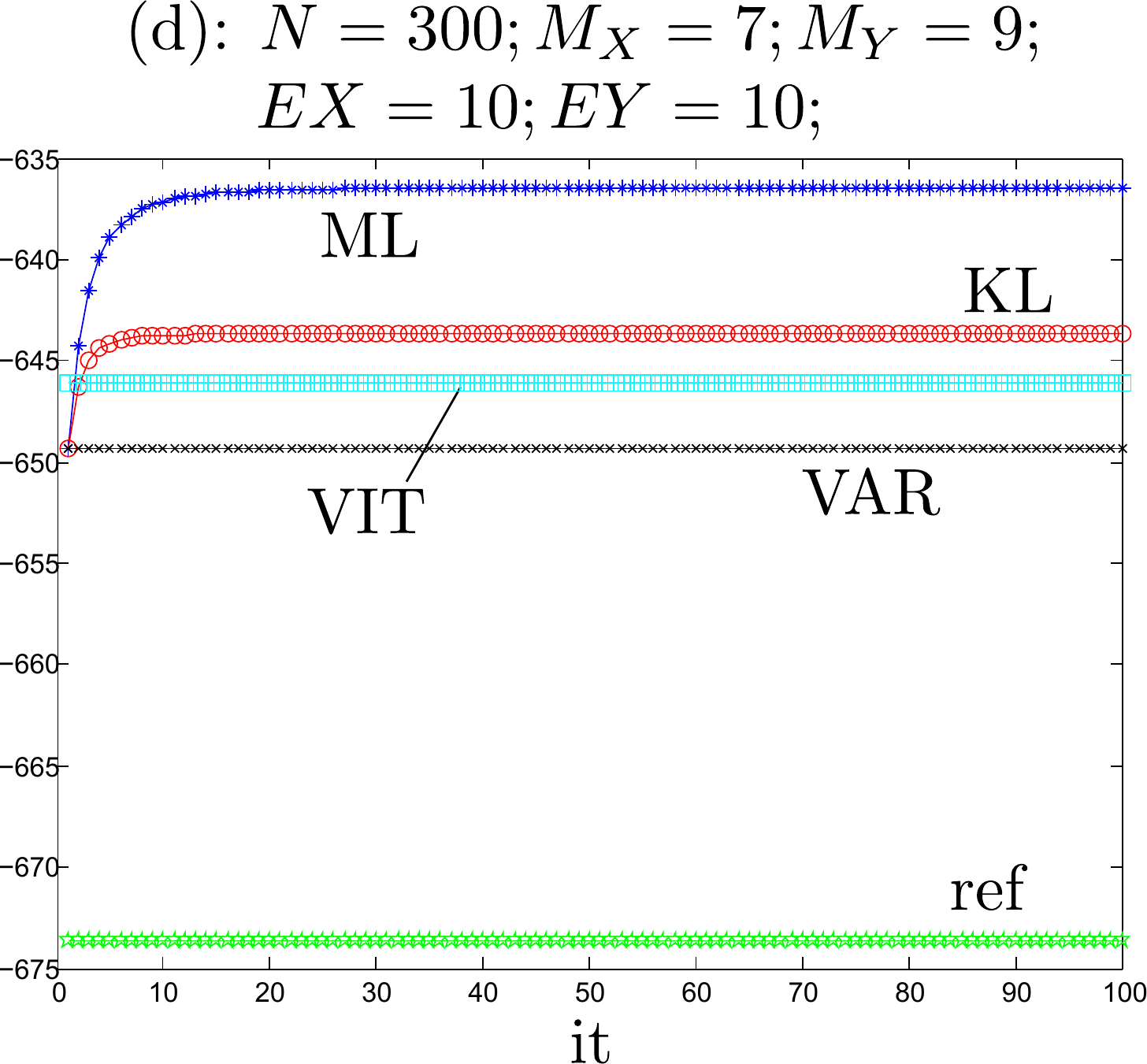}
\caption{Evolution of the log-likelihood for iterative learning of  a single block from randomly generated sets of pairs 
$\left\{ ({\bf f}_{X[n]}, {\bf b}_{Y[n]}), n=1:N \right\}$}
\label{fig:one_block_sim}
\end{figure}

\begin{figure}[ht]
\center
\includegraphics[width=7.0cm]{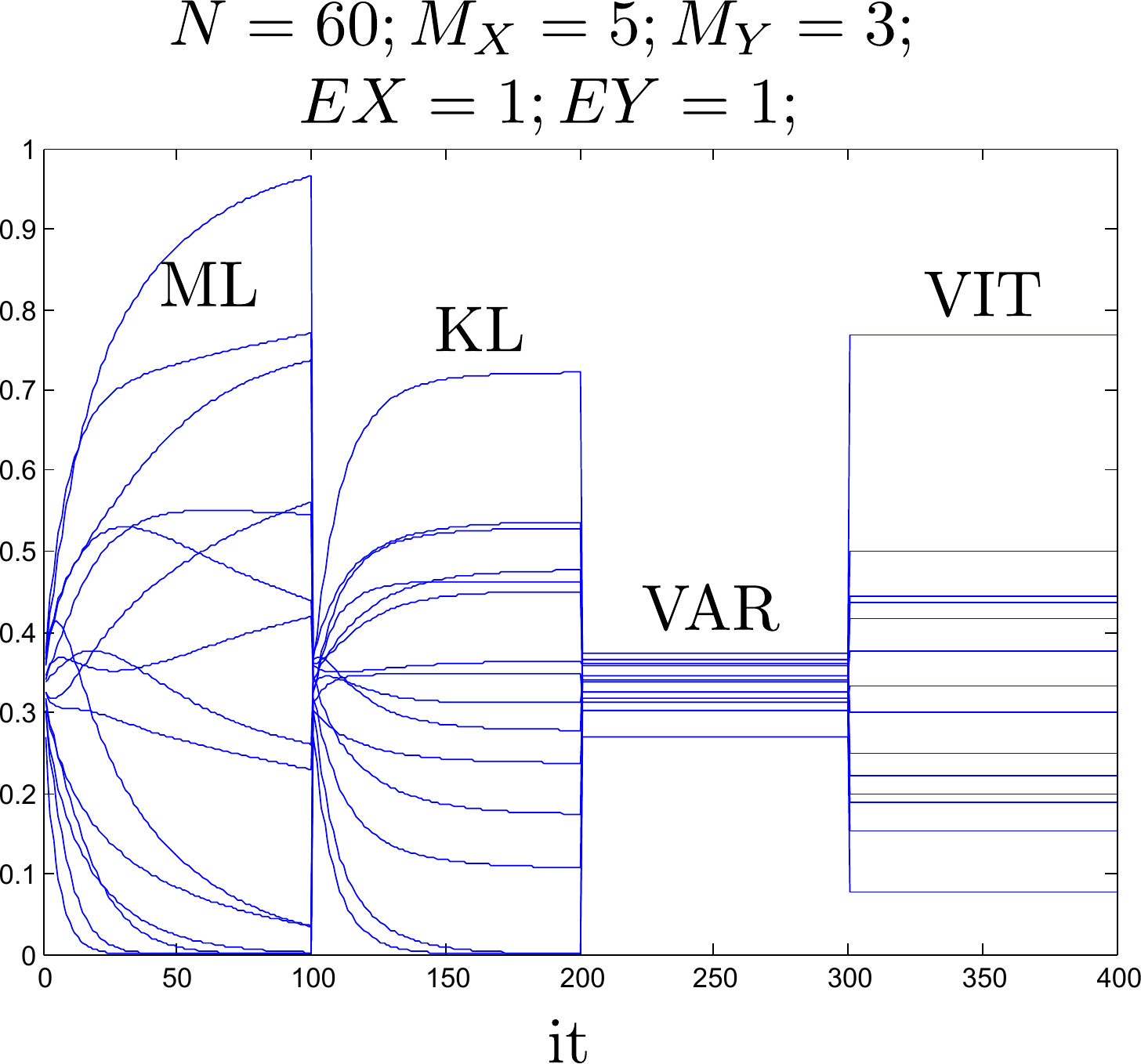}
\includegraphics[width=7.0cm]{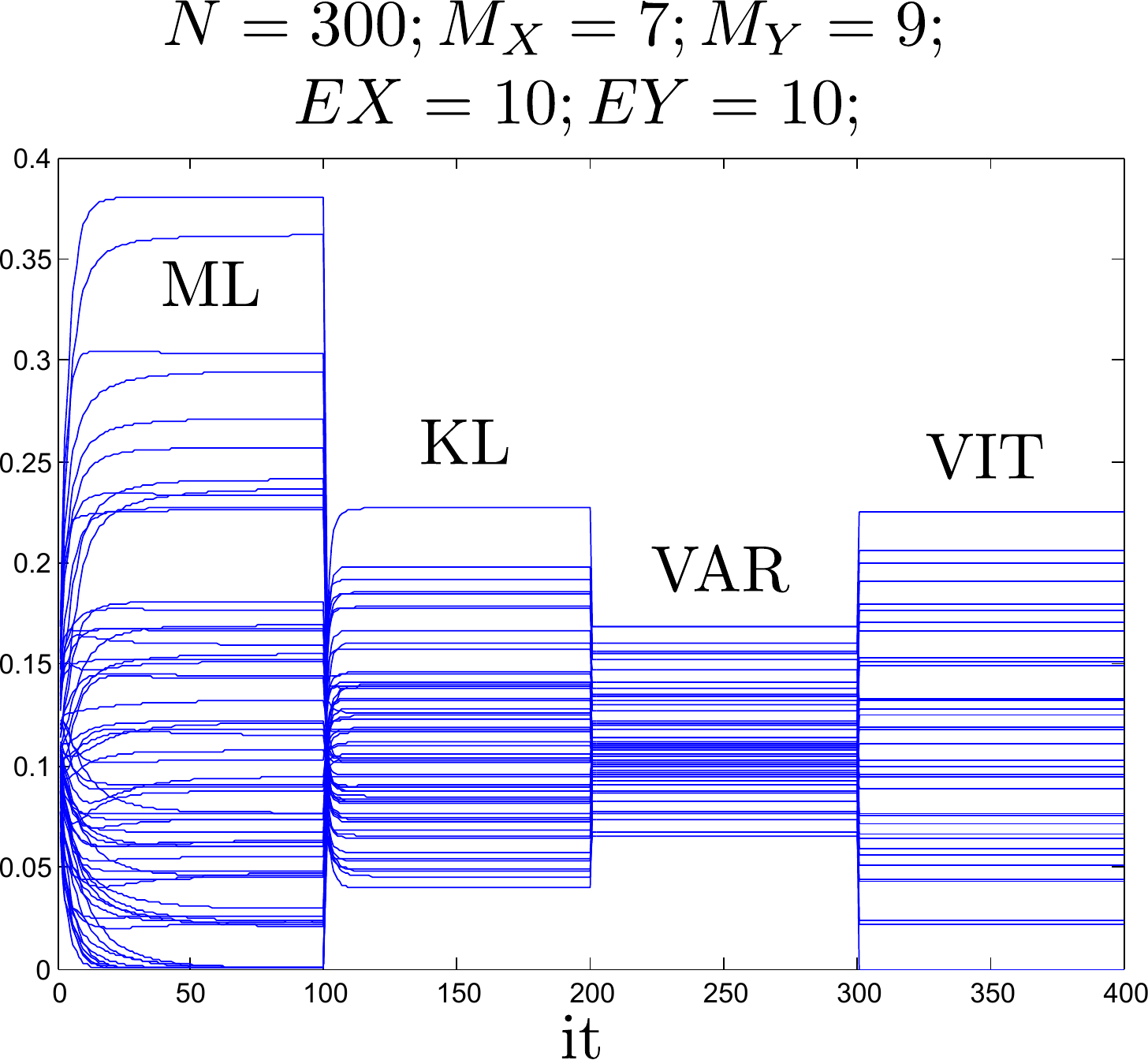} 
\caption{Evolution of the probabilities in $\theta$ for one-block learning  for ML and KL algorithms. The coefficients found with VAR and VIT algorithms (non recursive) are reported ot the same scale for comparison. }
\label{fig:one_block_coeff}
\end{figure}
In Figure \ref{fig:one_block_sim} we show the typical evolution of the log-likelihood $\sum_{n=1}^N   \log \left( {\bf f}_{X[n]}^T~ \theta~ {\bf b}_{Y[n]} \right)$, for various values of  $M_X$, $M_Y$, $N$, $EX$ and $EY$ during 100 iterations on the whole data set for the Maximum Likelihood (ML) and the Minimum Divergence (KL) algorithms. The Variational  (VAR) and the Viterbi-like (VIT) algorithms, that are  non recursive, are run one time and their log-likelihoods are shown on the same graph. We have also included the log-likelihood for a randomly generated  matrix (rows sampled from a uniform distribution) as an independent reference (ref). We see that, despite the similarities in the updating equations, the ML algorithm consistently provides the best result. The KL algorithm is usually the next best with the VIT algorithm providing most of the time a better solution than the VAR algorithm.  Note that when the distributions are sharpened,  as in Figures \ref{fig:one_block_sim}(b) and \ref{fig:one_block_sim}(d), all the algorithms, as expected,  tend to give very similar solutions. Further sharpening would  make the results essentially indistinguishable. 

Figure \ref{fig:one_block_coeff} shows also coefficient evolution for two cases. 
The ML and the KL algorithms show very smooth convergence with the ML algorithm spanning a larger range for coefficients in comparison to the other algorithms. The KL algorithm is generally faster than the ML. This is due to the fact that the probability values in ML compete with the whole matrix during adaptation, while in KL the competition is only column-wise. 
 
Simulations for checking generalization  were also performed by splitting the dataset in two halves. The first one for training and the second one for testing. 
The results are essentially indistinguishable from the ones presented above as far as the number of examples $N$ is large enough. This is not surprising since the data is generated in stationary mode and the cardinalities are relatively small.

\section{Simulations on Larger  Architectures}
\label{sec:larger}
To compare the performance of different algorithms we have to face various difficulties because the available data, typically in form of instantiated variables, are generally not 
connected to a specific architecture, but they are the result of measurements or inferences coming from unknown sources. Parameter learning has to discover the hidden relationships among the variables and this is usually done by guessing an architecture and performing training on it. 
Much of the results depend on the specific architecture and the hidden variables' cardinalities. The best we can do is to compute a measure of the resulting  log-likelihood. To maintain better control on the data,  in the following we run simulations on two synthetic architectures where we first generate the data and then  do parameter learning. We experiment with various iteration numbers and cardinalities and with mismatching between the trained graph and the known ``true" generative model.

\subsection{One Embedding Variable}

In the first set of experiments we have simulated the architecture of Figure \ref{fig:exx1}
both in generative and in learning mode. Our objective is to compare the four 
algorithms in learning priors and conditional probabilities when we have a single embedding variable $S$. 

We have generate independent samples $\{(X_1[n]X_2[n]X_3[n]), n=1:N\}$ with $N=400$ and the following parameters: $ M_S=4$; $M_{X_1}=2$; $M_{X_2}=2$; $M_{X_3}=3$;
$\Pi (S)=[0.25~0.25~0.25~0.25]$; 
$P(X_1|S)=[0.1~0.9;0.1~0.9;0.9$ $0.1;0.3~0.7]$;
$P(X_2|S)=[0.1~0.9;0.99$ $0.01;0.5~0.5;0.2~0.8]$;
$P(X_3|S)=[0.1~0.89$ $0.01;0.3~0.3$ $0.4 ;0.8~0.1~0.1;0.1~0.8~0.1]$. The samples are generated  as delta distributions at the source $S^{(0)}$ and propagated in the diverter, where they are fed to the conditional probability blocks where again delta  distributions are randomly generated for $X_1$, $X_2$ and $X_3$. These samples are used as backward distributions in the learning phase. 

\begin{figure}[h]
\center
\includegraphics[width=7.0cm]{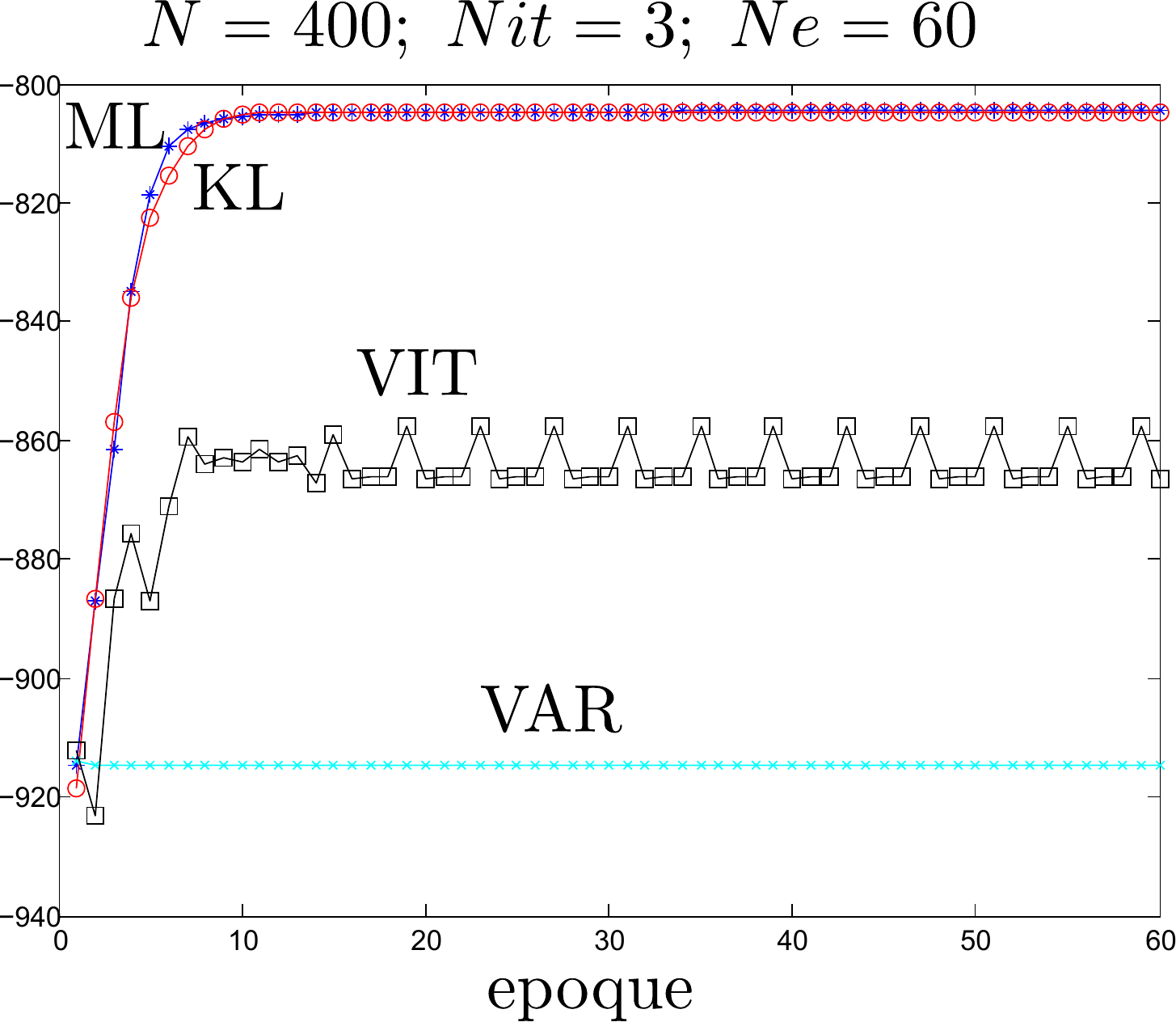}
\hspace{0.2 cm}
\includegraphics[width=7.0cm]{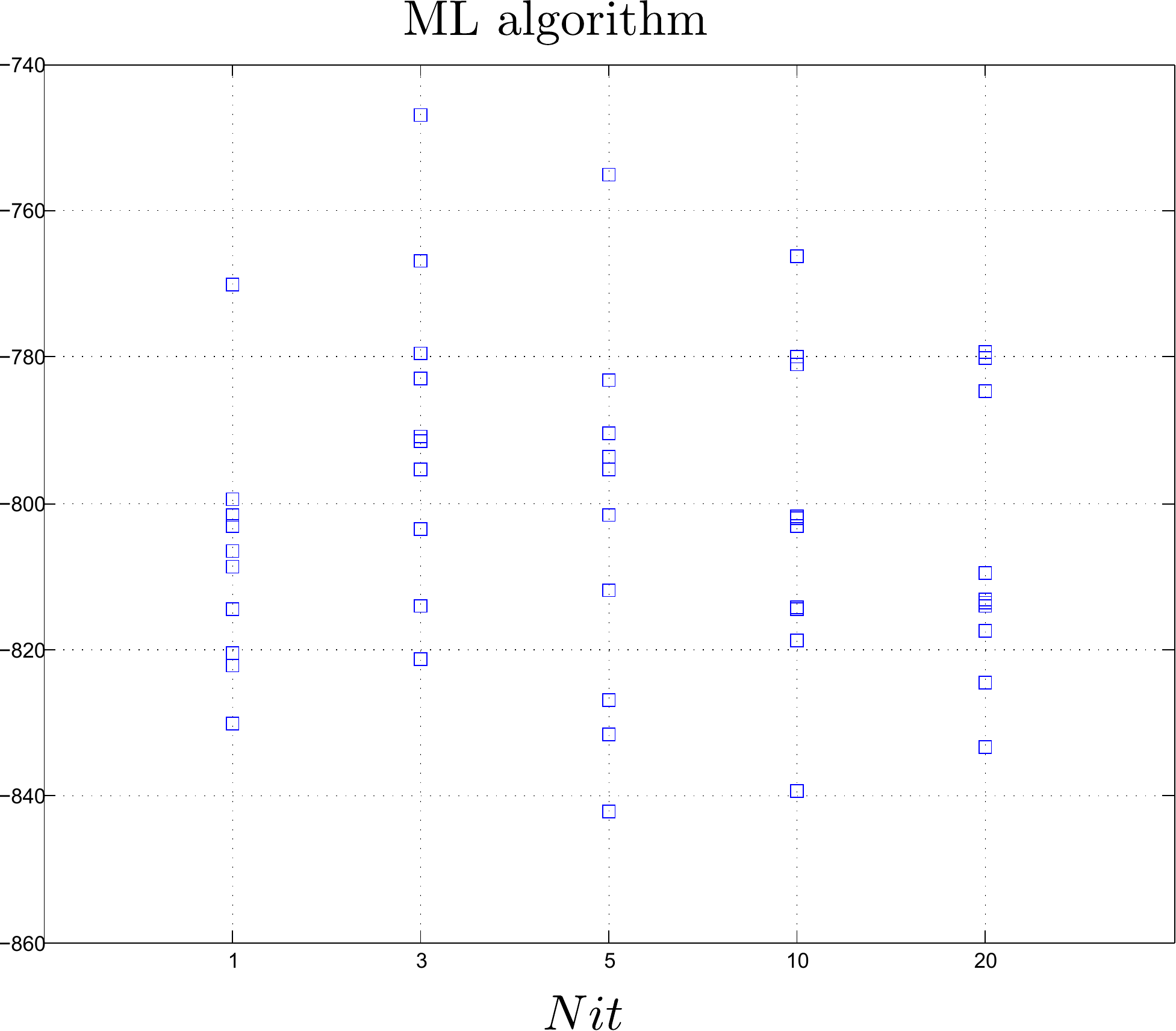} 
\caption{Left plot: Typical evolution of the log-likelihood for the four algorithms  for $Neq=60$  EM cycles. Right plot: Scatter diagram of the log-likelihood values obtained at convergence with the ML algorithm for  $Nit=1,3,5,10,20$.   }
\label{fig:one_emb_l}
\end{figure}

\begin{figure}[h]
\center
\includegraphics[width=7.0cm]{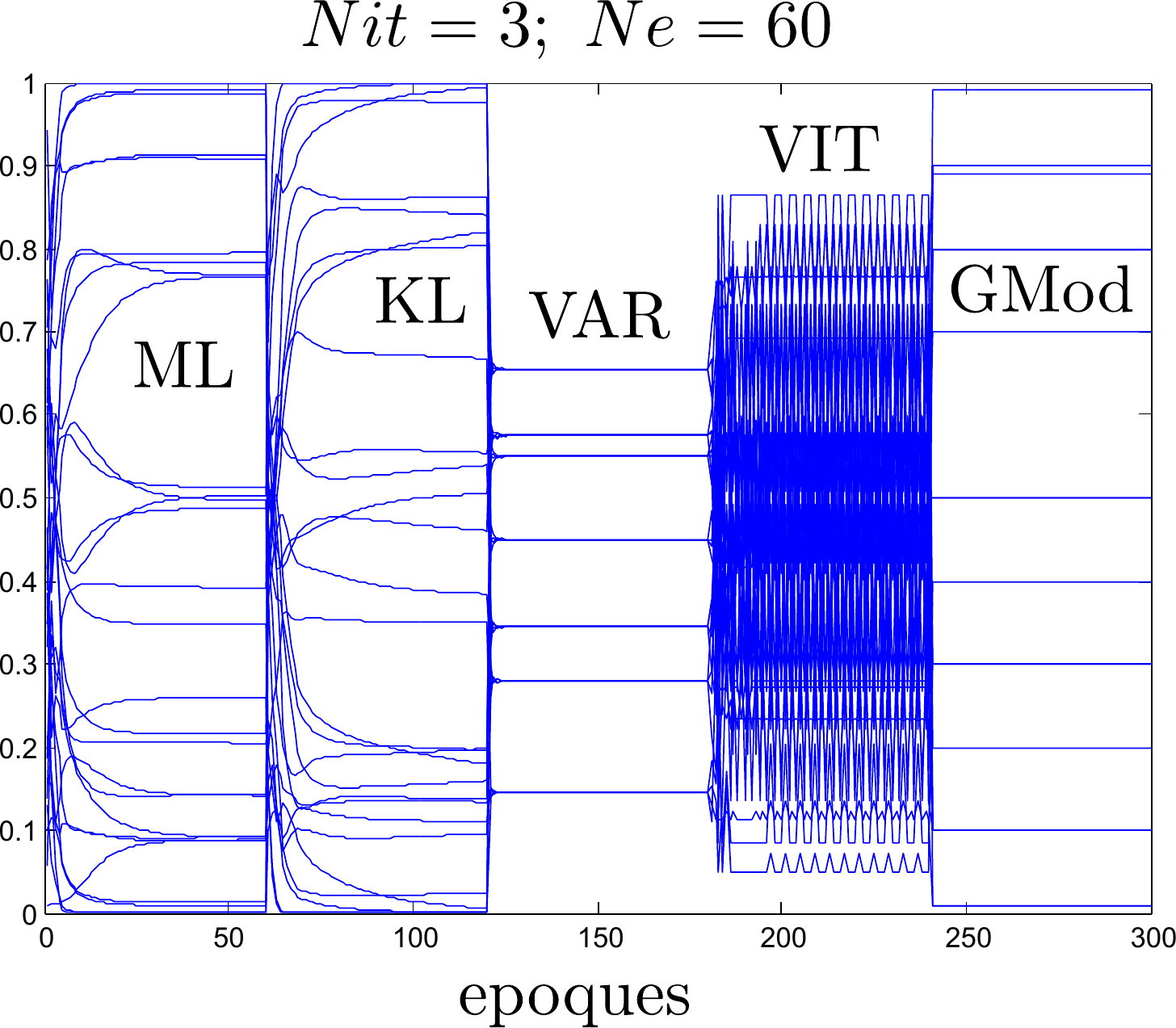}
\includegraphics[width=7.0cm]{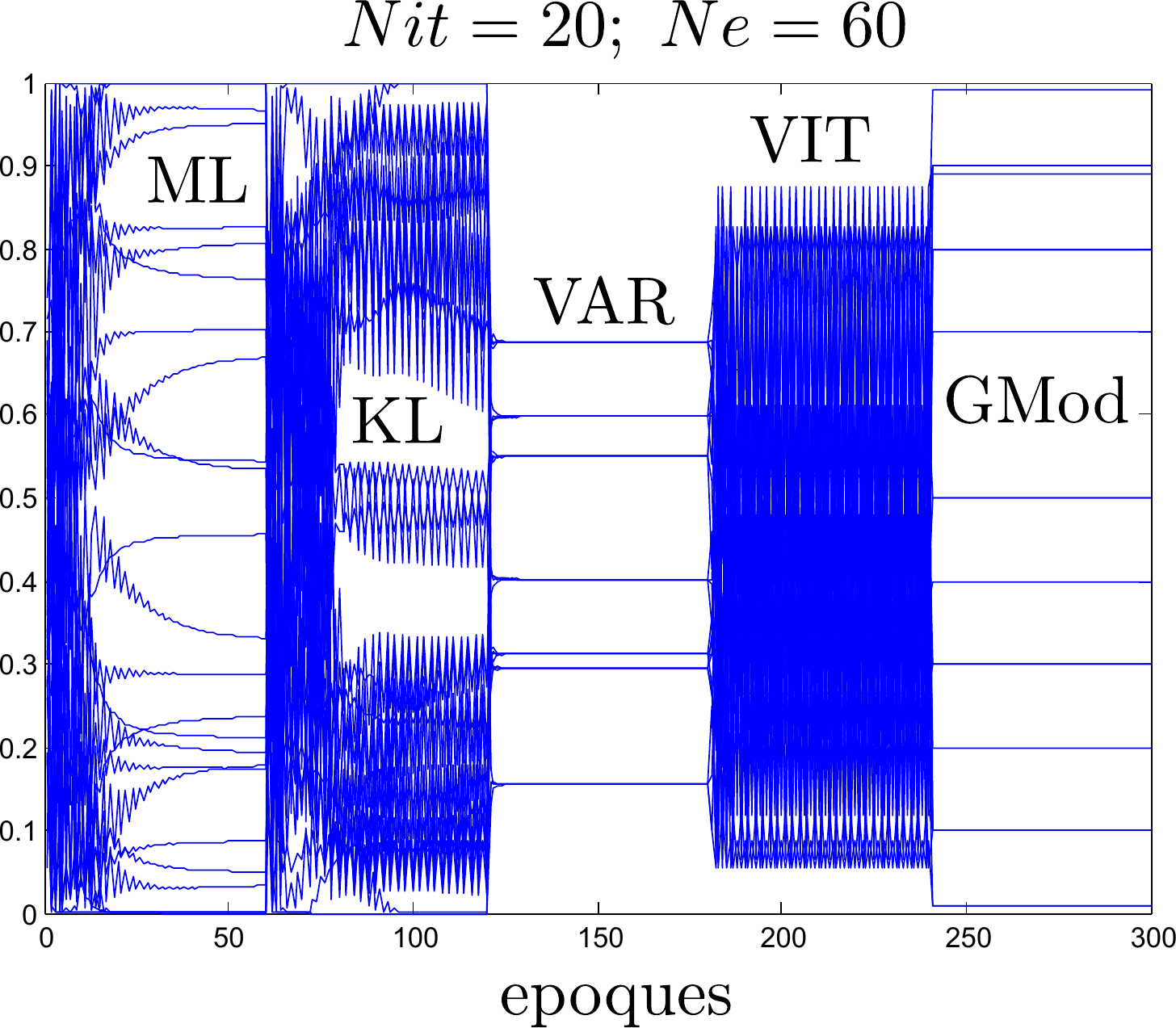} 
\caption{Typical evolution of the probabilities in $P(X_1|S)$, $P(X_2|S)$, $P(X_3|S)$ for 60 epoque cycles for the four algorithms and for two different values of $Nit$. The fifth column in the two graphs reports on the same scale the values used in the generative model (GModel).}
\label{fig:one_emb_coeff}
\end{figure}
\noindent
{\em Experiment 1:} For learning the parameters of the graph, in this initial experiment,  we have set  the learning mask $L[n]$  to one for all $n$ including   the same $N=400$ data points both in training and in testing. The graph on which we perform learning has the same structure and cardinalities as the generative model ($M_S=4$). The whole data set is presented to the system  $Ne$ times, each time for an EM cycle (epoque).  The two recursive algorithms (ML and KL) are run for $Nit$ iterations for each epoque. All the messages in the graph are initialized to independent uniform random values in $[0~1]$. 
Results of learning are evaluated by computing at the end of each epoque the aggregated log-likelihood value 
\begin{equation}    
l=\sum_{i=1}^3\sum_{n=1}^N \log[{\bf 1}_{M_{X_i}}^T ({\bf f}_{X_i[n]} \odot {\bf b}_{X_i[n]} )].
\label{eq:aggl}
\end{equation}
All messages are kept normalized during propagation and for log-likelihood evaluation. 
Figure \ref{fig:one_emb_l} (left)  shows a typical evolution of the log-likelihood  as the epoques progress from 1 to $Ne=60$ for $Nit=3$. The best performances are  obtained with both the ML and the KL algorithms that show very similar results. The VIT algorithm gives consistently an inferior  performance while the VAR algorithm does not seem to provide acceptable results. We attibute this failure to the fact that the variational approximation is based on smooth averages that do not allow sufficient differentiation among the various realizations.  

The coefficients for the priors in $\Pi_S$ tend to become uniform in all cases and are not shown here for brevity. 

To evaluate the effects of the the number of iterations $Nit$ per epoque  for the iterative algorithms ML and KL,  we have repeated each simulation 10 times with different random initial conditions for  various values  of $Nit=1, 3, 5, 10,20$. The final log-likelihoods are shown in Figure \ref{fig:one_emb_l} (right) only for the ML algorithm (the KL algorithm gives very similar values). It is clear that after each run the algorithm reaches one of the many local minima. We see also that not much difference in the quality of the solution is obtained by changing $Nit$. There is a certain (weak) preference for $Nit=3$. It is therefore advisable to keep $Nit$ small to contain  the computational complexity.  In any case large values of $Nit$ render the algorithm ``too greedy" in reaching good temporary solutions that need large changes in subsequent EM cycles, as seen also in the following coefficient analysis.   

Typical coefficient evolutions are 
shown in the columns of Figure \ref{fig:one_emb_coeff} for the four algorithms  $Nit=3$ (left), $Nit=20$ (right). The last column in the two figures represents the generative model parameters (constant on the same scale). From these plots we can derive the following considerations: a) the solutions reached in the various algorithms are different from the true values used in the generative model even if the log-likelihoods are very similar (this is an identifiability issue that should be more specifically addressed, but it goes beyond the scope of this paper); b) the iterative algoritms ML and KL span a range for the parameters which is wider than the one obtained with the VAR and the VIT algorithms; c) for $Nit=3$ both the ML and the KL algorithms show nice smooth convergence, while the VIT algorithm shows wide oscillations; c) for $Nit=20$ also ML and KL algorithms show an oscillating behavior before convergence, as expected, because of the greedy nature of the updates; d) the log-likelihoods converges at a speed faster than the coefficients (this is common in adaptive filtering where the cost function typically converges before the coefficients).  

\begin{figure}[h]
\center
\includegraphics[width=7.0cm]{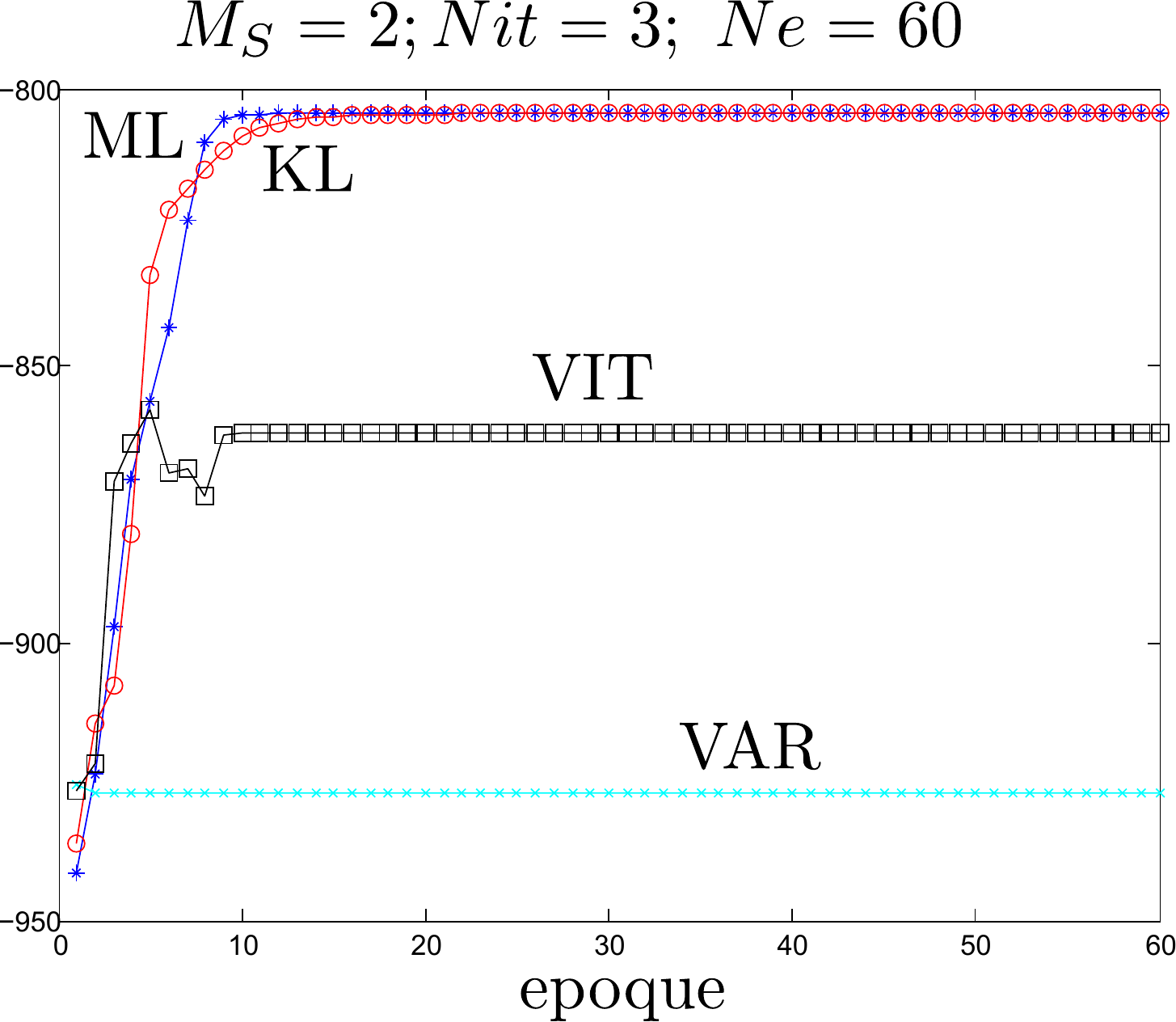}
\hspace{0.2 cm}
\includegraphics[width=7.0cm]{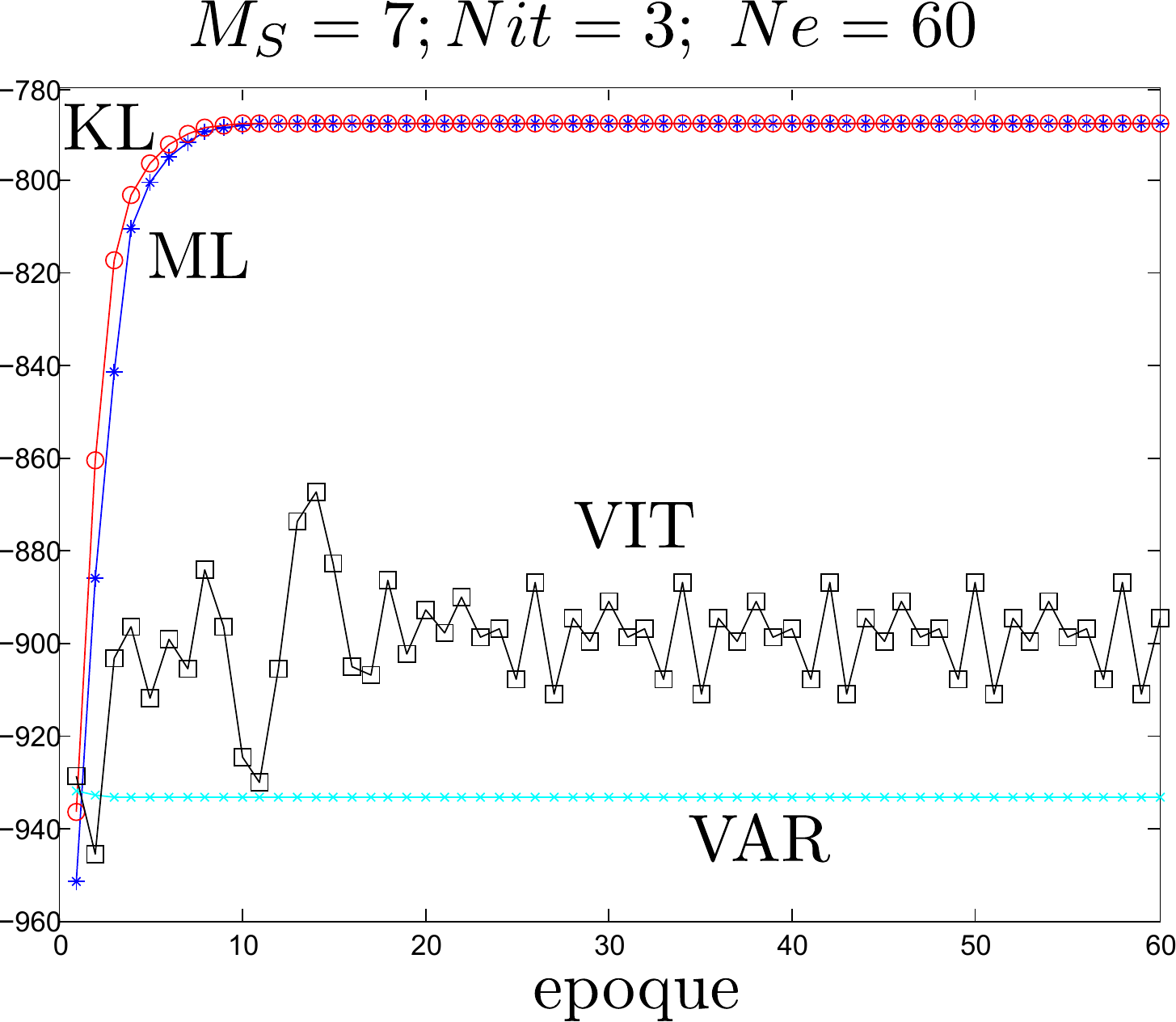} 
\caption{Test for Mismatch. Left plot: Typical evolution of the log-likelihood for the four algorithms  for $Neq=60$  EM cycles, $Nit=3$ and $M_S=2$. Right plot: Typical evolution of the log-likelihood for the four algorithms  for $Neq=60$  EM cycles, $Nit=3$ and $M_S=7$.   }
\label{fig:one_emb_l_mis}
\end{figure}

\noindent
{\em Experiment 2}: Test for mismatch between graph and generative model. 

 In this second set of experiments we have mismatched the size of the embedding space with respect to the generative model. We have used the same parameters used in Experiment 1, but with $M_S=2$ and $M_S=7$. The convergence behaviors are very similar to the ones obtained in Experiment 1 and are shown in Figures \ref{fig:one_emb_l_mis}. The  final log-likelihoods  are clearly worse than the ones obtained for $M_S=4$ when $M_S=2$.  The coefficient evolutions show behaviors similar to Figure \ref{fig:one_emb_coeff}(left) and are not shown for brevity

 \noindent
{\em Experiment 3}: Test for generalization. 
\begin{figure}[h]
\center
\includegraphics[width=7.0cm]{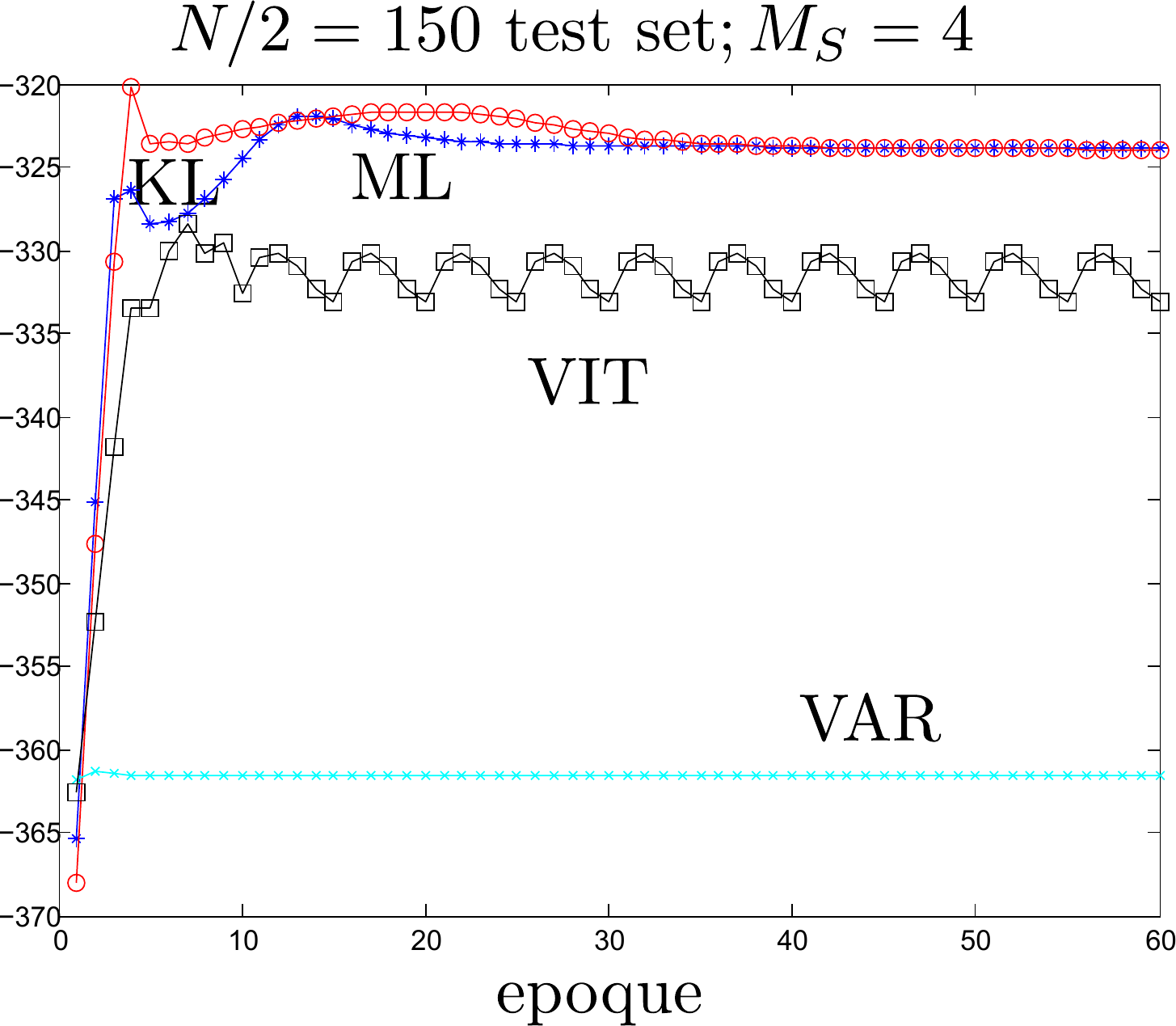}
\hspace{0.2 cm}
\includegraphics[width=7.0cm]{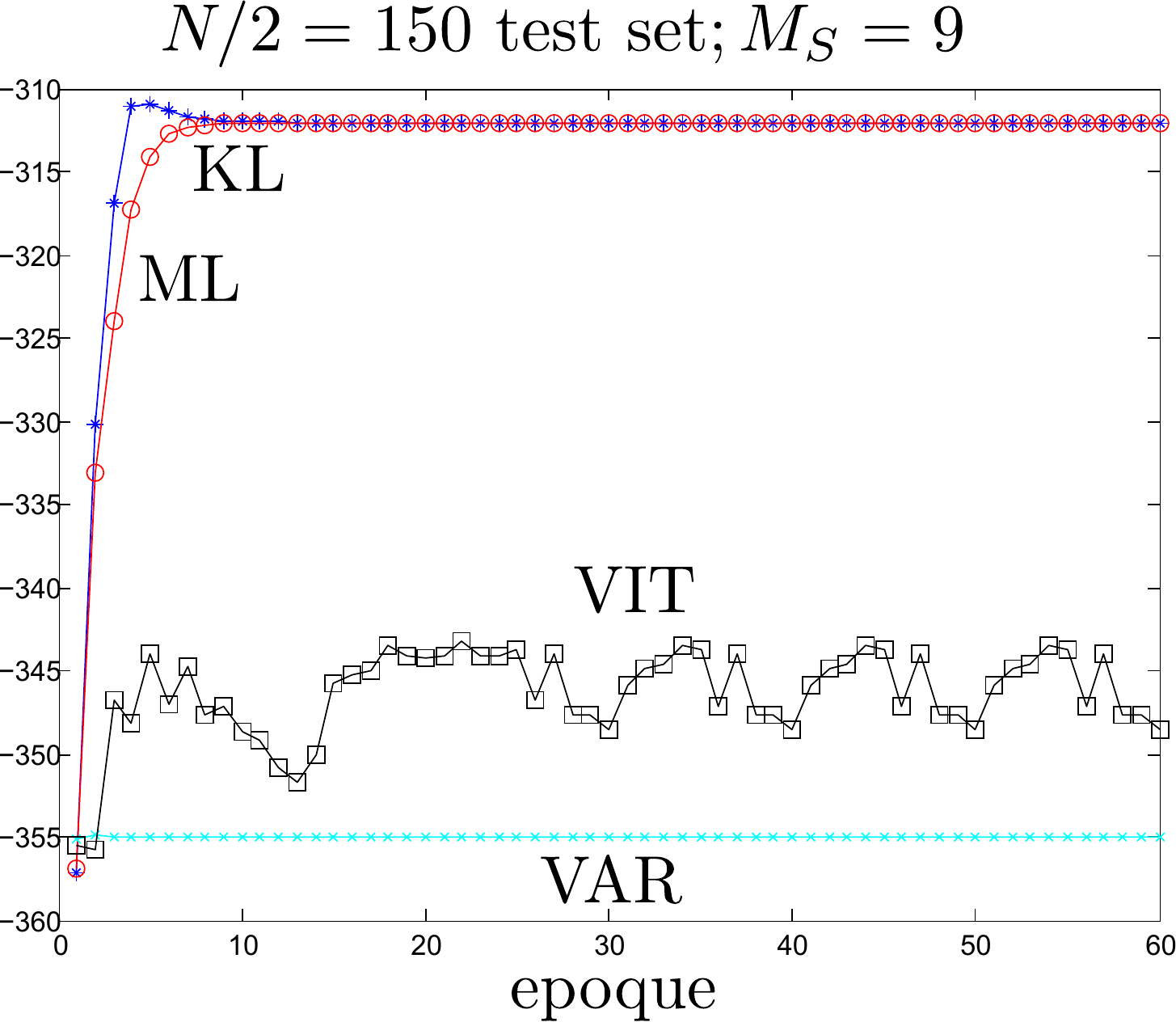} 
\caption{Test for generalization: typical evolution on the test set of the log-likelihood for the four algorithms  for $N=300$ $Neq=60$  EM cycles, $Nit=3$ and $M_S=4$ (left plot), 
$M_S=9$ (right plot). In this experiment the first $N/2$ samples are used for training and the remaining $N/2$ for testing. }
\label{fig:one_emb_gen}
\end{figure}

In this set  of experiments we have split  $N=300$ samples in two halves. The first part is used for training and the second one for testing. Figure  \ref{fig:one_emb_gen} shows the evolution evaluated on the second half (test set) while training is on the first half (trainig set) for $M_S=4$ and for $M_S=9$.   The evolutions are quite smooth showing how 150 samples are sufficient to capture stationarity in the data. Similar results are obtained with different parameters and data sets.  

\subsection{A Deeper Graph}
\begin{figure}[ht]
\center
\includegraphics[width=6.5cm]{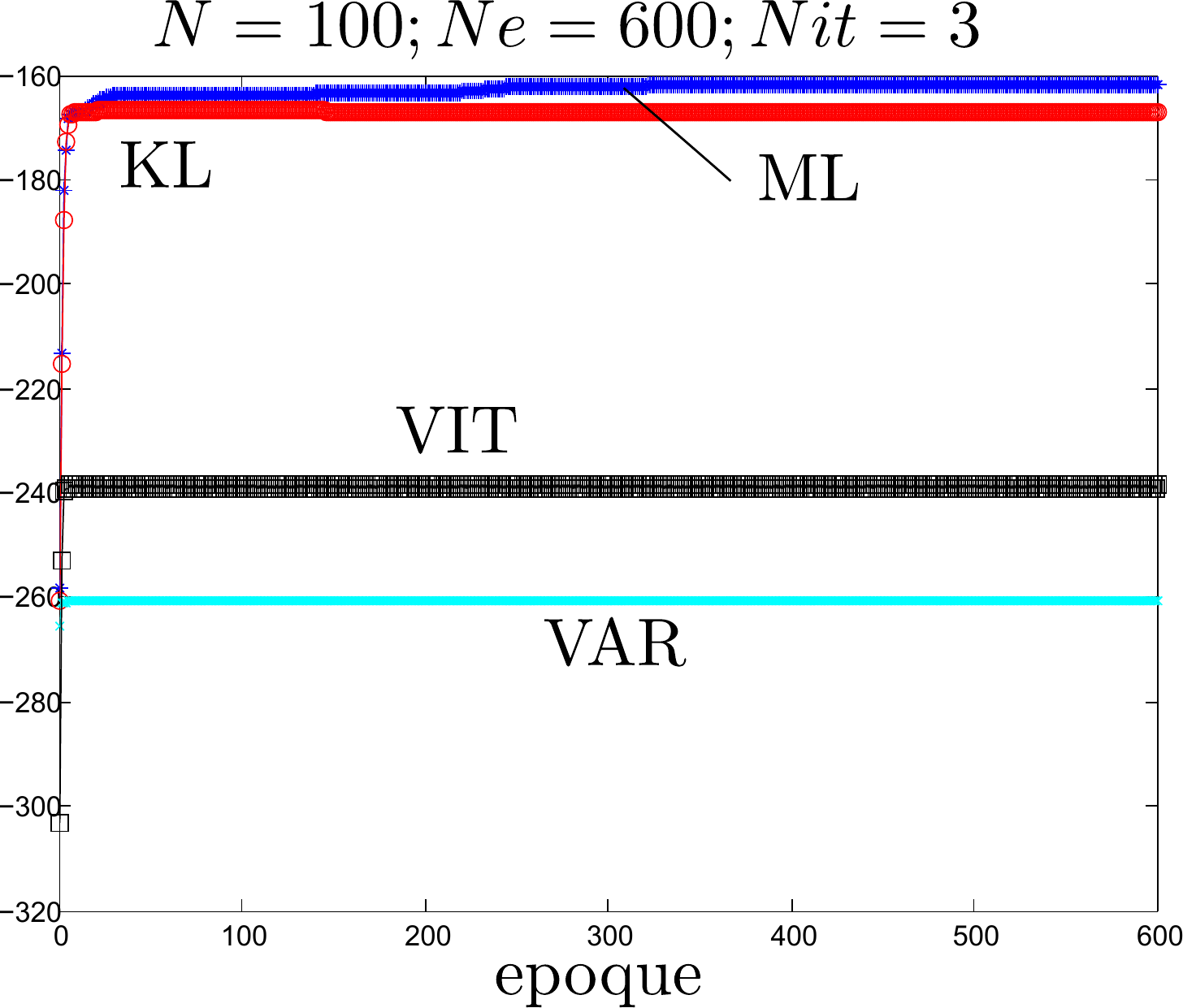}
\includegraphics[width=6.5cm]{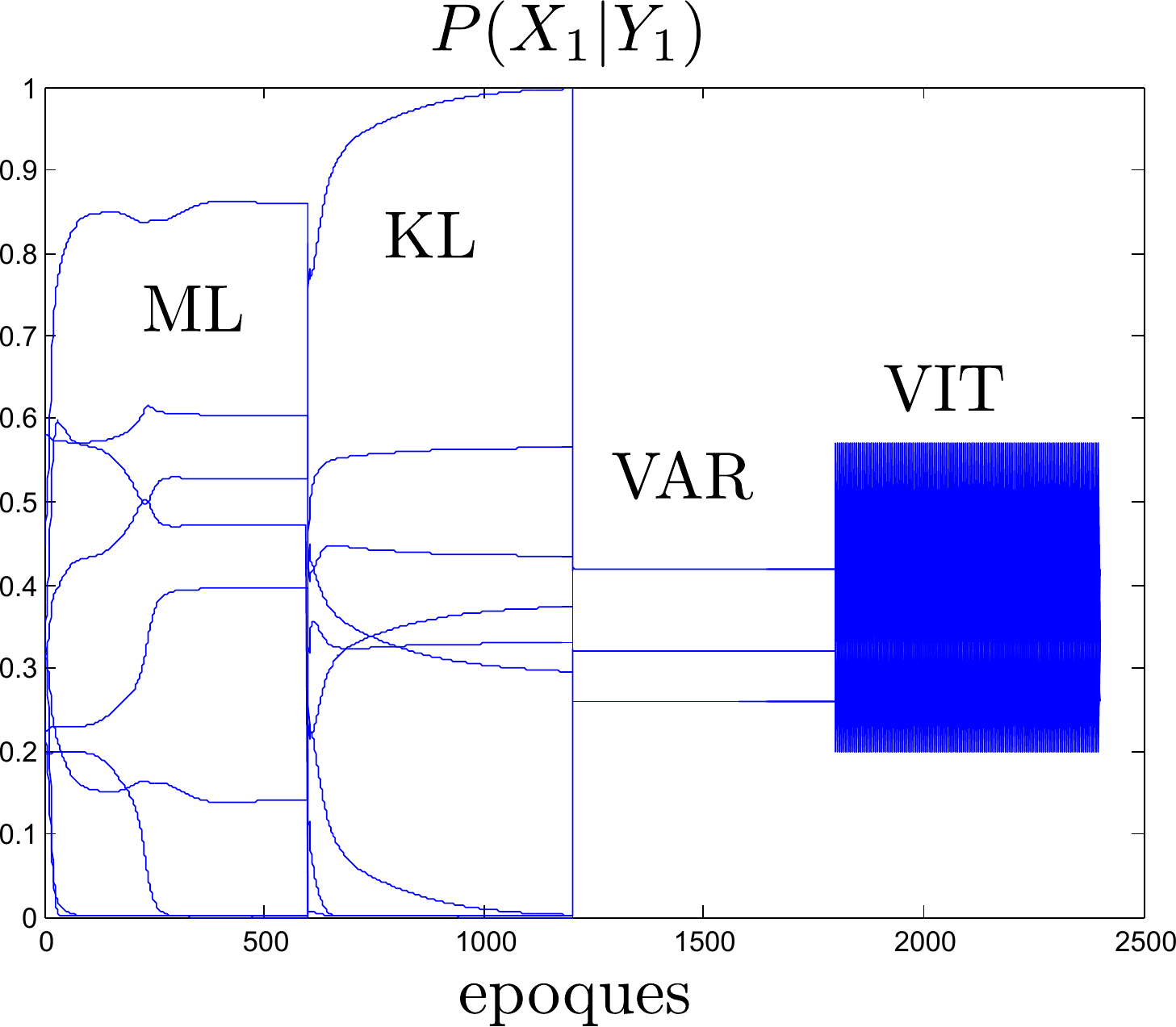} 
 \includegraphics[width=6.5cm]{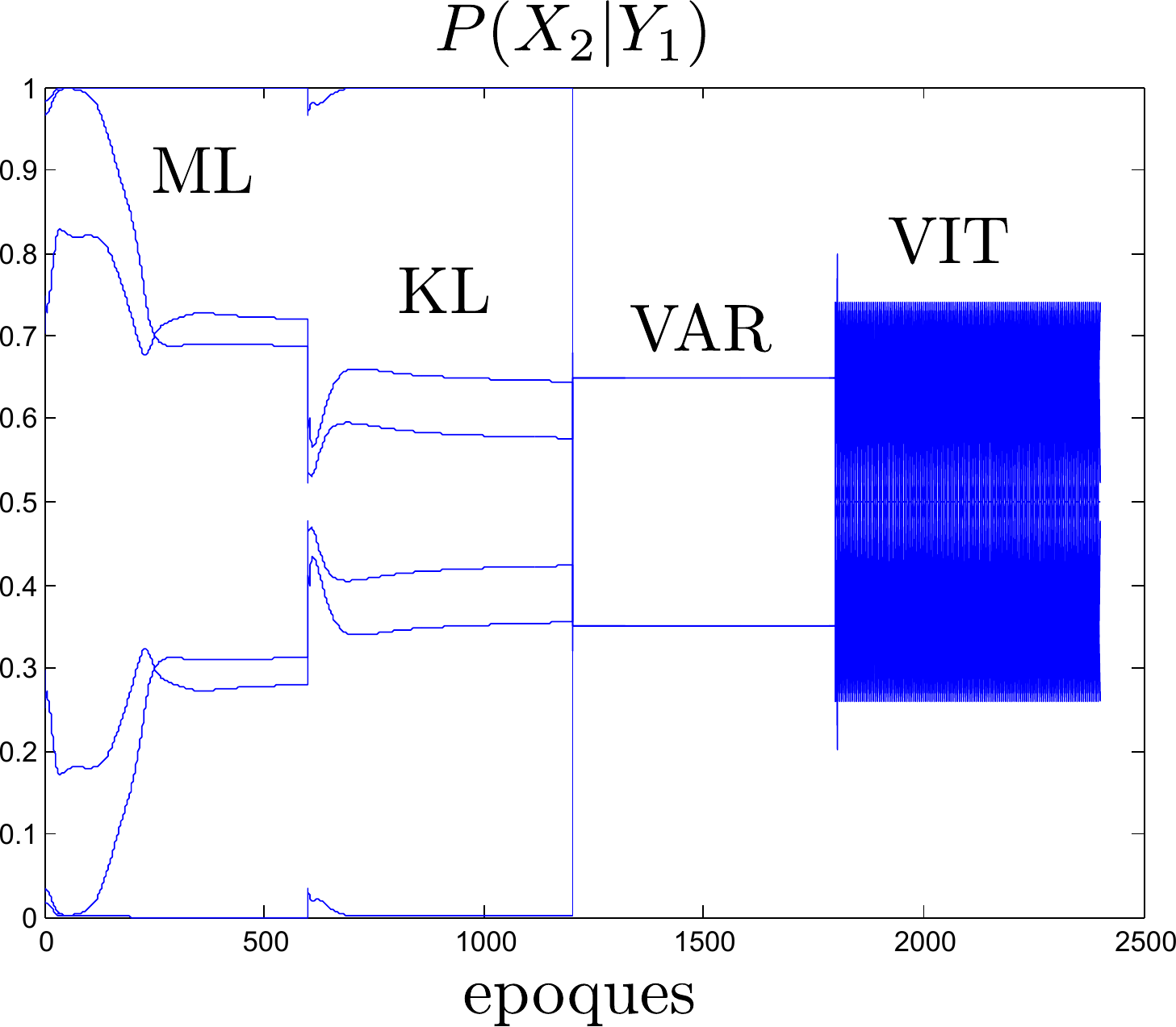}
\includegraphics[width=6.5cm]{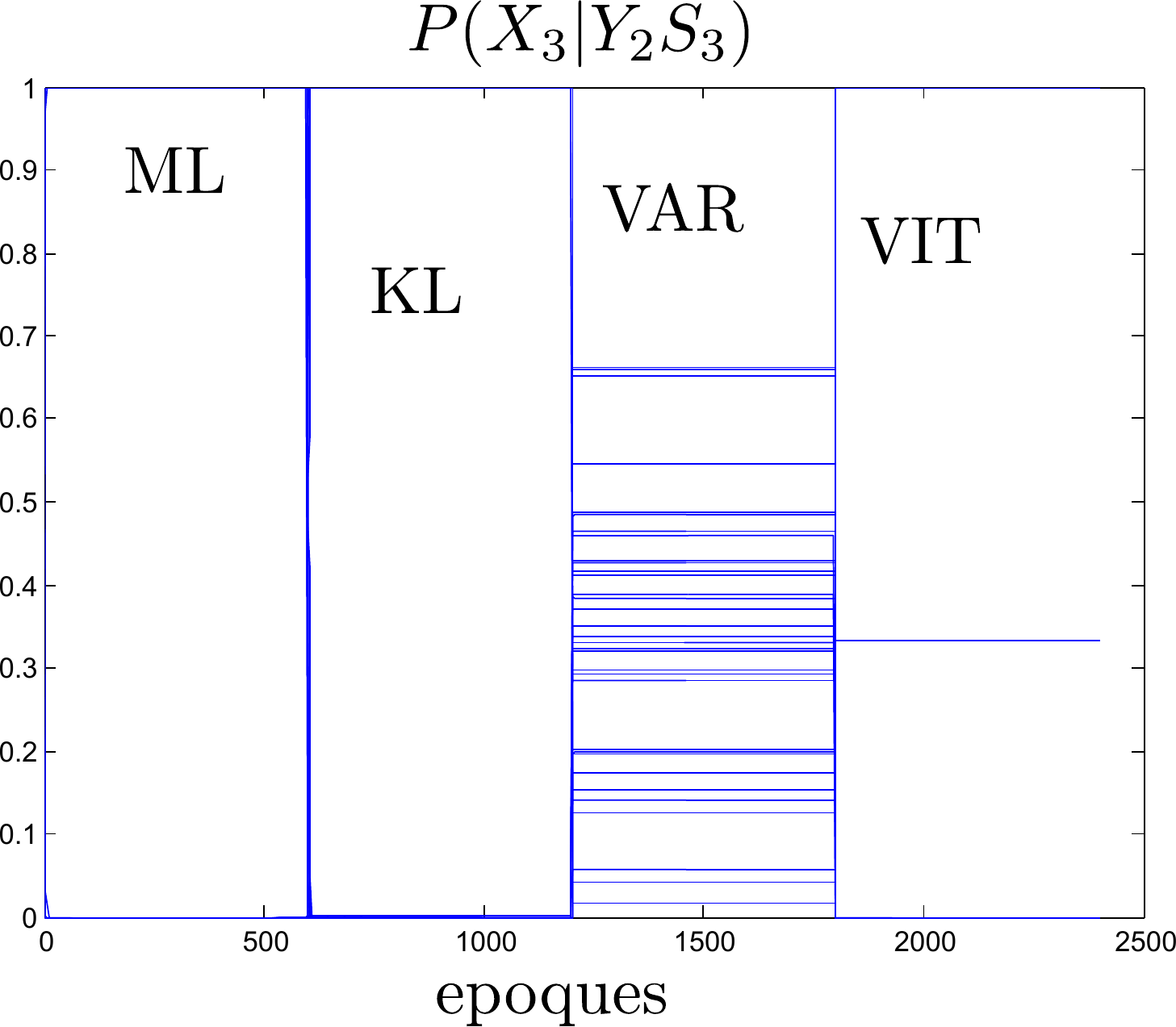}
\includegraphics[width=6.5cm]{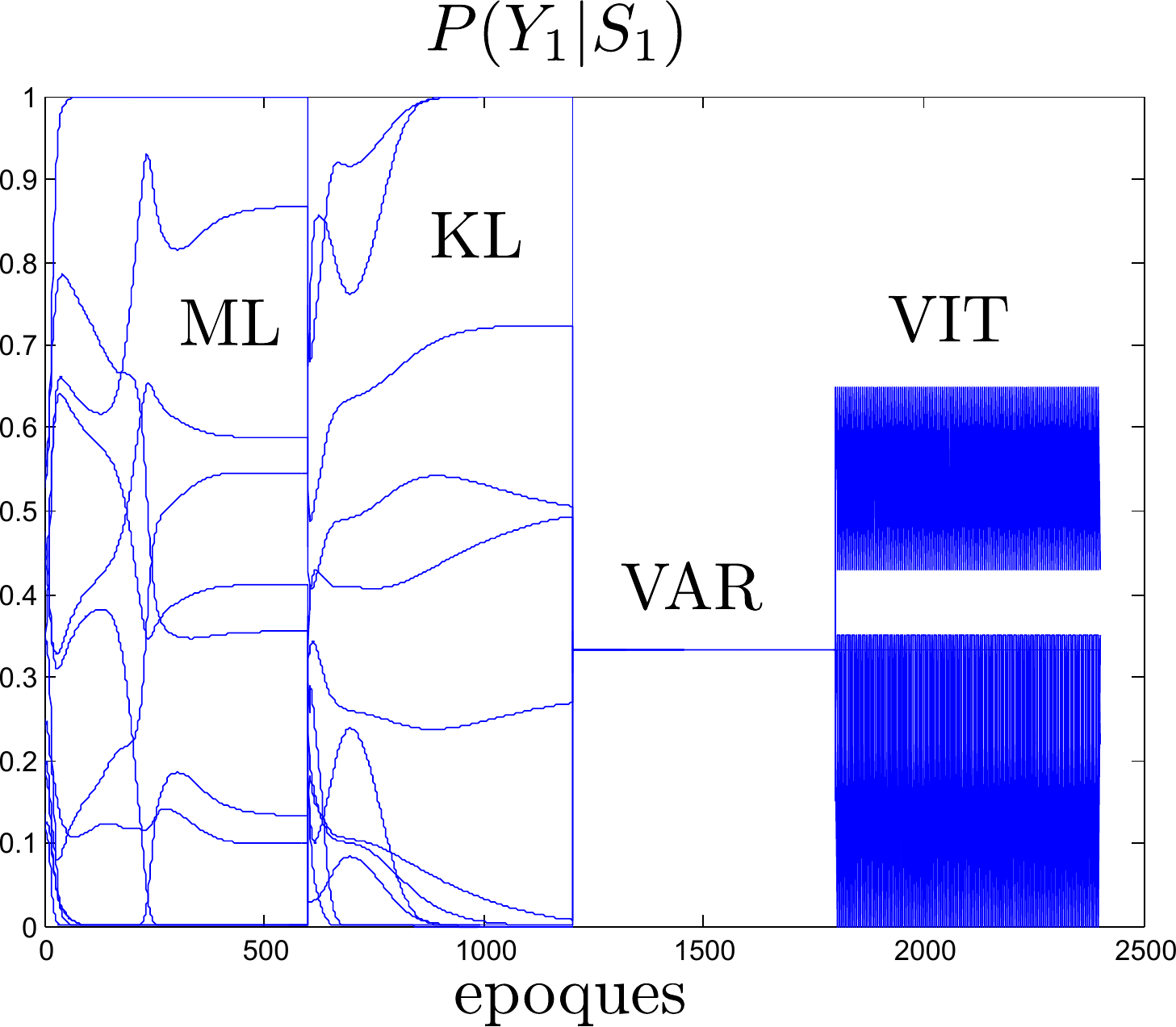} 
\includegraphics[width=6.5cm]{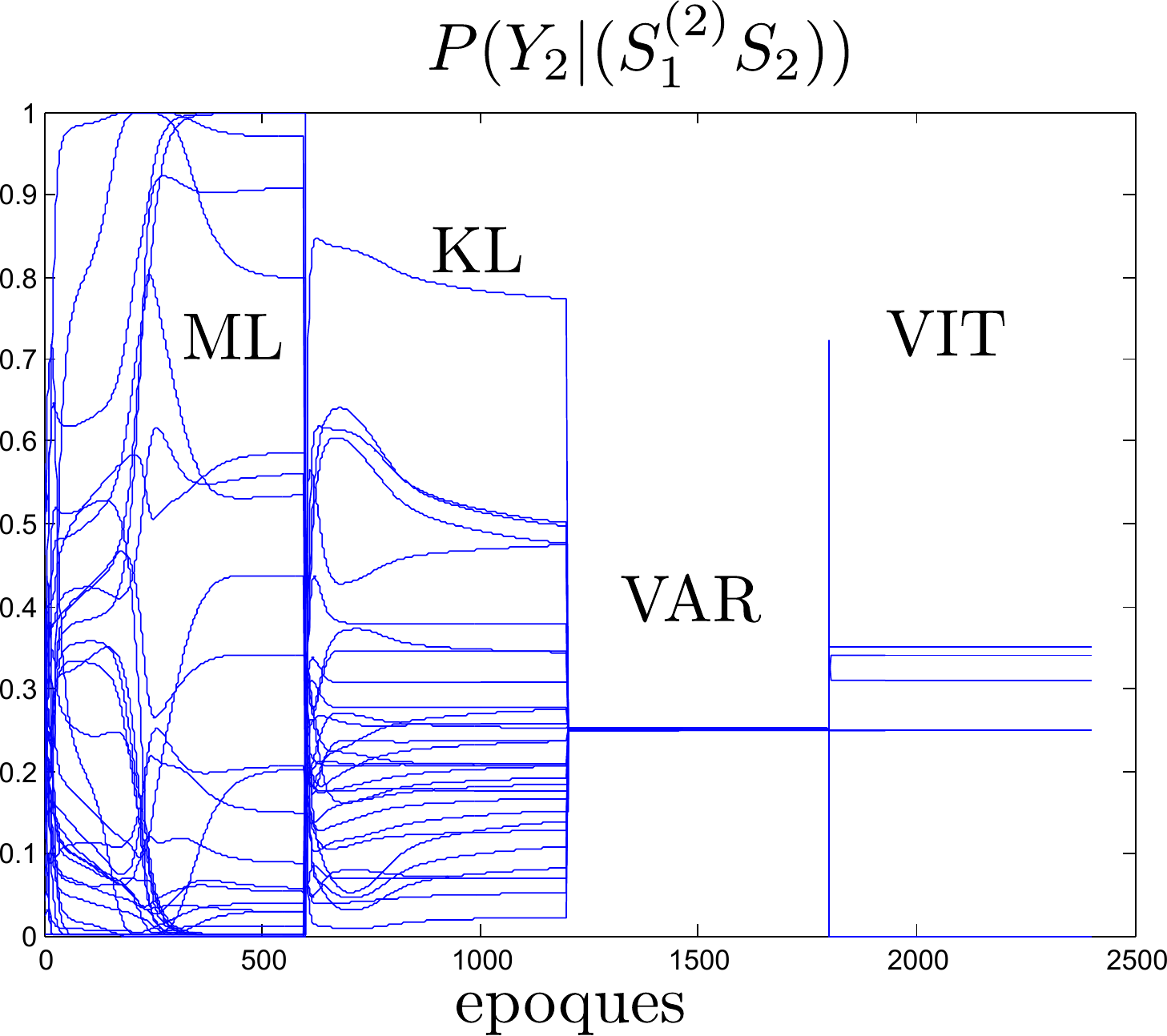}
\caption{Results of the simulations on the graph of Figure \ref{fig:exx4}. The first plot (top left) shows the evolution of the aggregated likelihood on $Ne=600$ EM cycles. The other plots show the evolution of the coefficients in the other  blocks.}
\label{fig:arch}
\end{figure}

In this section we present the results of simulation on the larger graph of Figure \ref{fig:exx4} where some of the learning blocks are more deeply embedded in the architecture. We have set the parameters to: $M_{S_1}=4$; $M_{S_2}=2$; $M_{S_3}=3$; $M_{Y_1}=3$; $M_{Y_2}=4$;
$M_{X_1}=3$; $M_{X_2}=2$; $M_{X_3}=3$. The shaded blocks are fixed to the matrices given in formulas (\ref{eq:margt}) in Appendix A:
\begin{equation}
\begin{array}{l}
P((S_1^{(2)}S_2)^{(1)}|S_1^{(2)})={ 1 \over M_{S_2}} I_{M_{S_1}} \otimes {\bf 1}_{M_{S_2}}^T; \\
P((S_1^{(2)}S_2)^{(2)}|S_2)={ 1 \over M_{S_1}} {\bf 1}_{M_{S_1}}^T \otimes  I_{M_{S_2}}  ; \\
P((Y_2 S_3)^{(1)}|Y_2)={ 1 \over M_{S_3}} I_{M_{Y_2}} \otimes {\bf 1}_{M_{S_3}}^T; \\
P((Y_2 S_3)^{(2)}|S_3)={ 1 \over M_{Y_2}} {\bf 1}_{M_{Y_2}}^T \otimes I_{M_{S_3}}.
\end{array}
\end{equation}
The other blocks have to be learned from examples.
We have used the architecture first in generative mode with hand-picked random parameters to generate $N=100$ examples as delta distributions at $X_1$, $X_2$ and $X_3$. These examples  are then used as backward delta messages at the same terminations. The variables' cardinalities in the learning graph are kept at the same values as in the generative model.       
We have compared the performances in terms of aggregated log-likelihood (\ref{eq:aggl}) for $Ne=600$ EM learning cycles (epoques). The number of iterations per epoque $Nit$ for the ML and the KL algorithm have been fixed to 3.
Figure \ref{fig:arch} shows the typical results of a simulation. The log-likelihood (top left) converges quite quickly while the coefficients (other plots) take longer to reach a stable value. The graph show the convergence for both visible and hidden-variable blocks.
The prior blocks tend all to become uniform and are not shown for brevity. 
It is clear how the superiority of the ML and the KL algorithms holds also for this architecture with smoothest convergence for the coefficients and a better final log-likelihood. The VIT algorithm often produces wide oscillations in the coefficients.  
We have run many more simulations to test for generalization and with mismatched values between generative model and learning graph. The  results are very similar and are not reported for brevity.   

\subsection{Other architectures}

We have performed other simulations on larger architectures with various topologies and variables' cardinalities. We have also applied the same algorithms to experimental data with hierachical architectures. The results are very consistent with the behavior shown in the above  experiments and seem to hold for large classes of cycle-free graphs.   
We believe that the results of these simulations may  provide a useful guidance to the designers of learning graphs for the applications.

\section{Conclusions and Future Trends}
\label{sec:concl}

In this paper we have discussed the application of four different learning algorithms to SISO blocks as they are embedded in a normal-form Bayesian graph. 
The superiority of the recursive multiplicative updates of the ML and the Kl algorithms, derived from robustified cost functions and an optimization procedure, has been demonstrated with simulations on synthetic data. We believe that the comparisons  presented in this paper, that to our knowledge have never been reported in the literature, may be useful to the programmer that wants to rapidly deploy an adaptive Bayesian graph. Further papers will report more specifically the application of this framework to deep architectures and to experimental data.

%\acks{We would like to acknowledge ........ }


%\newpage

\appendix

\section*{Appendix A: Product Space Mappings}
\label{app:margex}

Suppose we have $D$  discrete variables $X_1$, $X_2$,...,$X_D$, belonging to the spaces
${\cal X}_1$, ${\cal X}_2$,...,$\cal{X}_D$, having cardinalities $M_1$, $M_2$,...,$M_D$, respectively. The maps from the product space ${\cal X}_1\times {\cal X}_2\times...\times \cal{X}_D$ to each variable are described by the matrices 
\begin{equation}
\begin{array}{l}
P(X_1|X_1X_2...X_D)=I_{M_1} \otimes  1_{M_2} \otimes 1_{M_3} \otimes ... \otimes  1_{M_D}, \\ 
P(X_2|X_1X_2...X_D)=1_{M_1} \otimes I_{M_2} \otimes  1_{M_3} \otimes ... \otimes 1_{M_D},  \\
.... \\
P(X_D|X_1X_2...X_D)= 1_{M_1} \otimes  1_{M_2} \otimes  1_{M_3} \otimes ... \otimes I_{M_D},
\end{array} 
\label{eq:marg1}
\end{equation}
where $I_Q$ denotes the $Q \times Q$ identity matrix, $1_Q$  the $Q$-dimensional column vector with all ones, and $\otimes$ the Kronecker product.
All the matrices contain mostly zeros, but they are row-stochastic since only one element per row is equal to one. 

For example, matrices (\ref{eq:marg1}) define the blocks at the bottom of  Figure \ref{fig:exx1}, if $D=3$ and if $S$ is the whole product space:  $S=(X_1X_2X_3)$.  

Vice versa the maps from each variable to the product space are the transposed normalized of (\ref{eq:marg1})
\begin{equation}
\begin{array}{l}
P((X_1X_2...X_D)^{(1)}|X_1)={M_1 \over \prod_{i=1}^D M_i }I_{M_1} \otimes  1_{M_2}^T \otimes 1_{M_3}^T \otimes ... \otimes  1_{M_D}^T, \\ 
P((X_1X_2...X_D)^{(2)}|X_2)={M_2 \over \prod_{i=1}^D M_i }1_{M_1}^T \otimes I_{M_2} \otimes  1_{M_3}^T \otimes ... \otimes 1_{M_D}^T,  \\
.... \\
P((X_1X_2...X_D)^{(D)}|X_D)={M_D \over \prod_{i=1}^D M_i } 1_{M_1}^T \otimes  1_{M_2}^T \otimes  1_{M_3}^T \otimes ... \otimes I_{M_D}.
\end{array} 
\label{eq:margt}
\end{equation}
The matrices  are row-stochastic and have ${ \prod_{i=1}^D M_i \over M_j}$ non zero equal elements in each row, reflecting the uniform ambiguity that, given a single variable $X_j$, is left  on the product space. In the graph, the ambiguity is resolved around the diverter block by the intersection (product rule) with messages coming from the other variables.

For example, matrices (\ref{eq:margt}) can be applied to define the blocks that, in Example 2 of Figure \ref{fig:exx2}, go from $X_1$, $X_2$ and $X_3$ to $(X_1X_2X_3)^{(1)}$,  $(X_1X_2X_3)^{(2)}$ and $(X_1X_2X_3)^{(3)}$. Similarly matrices (\ref{eq:margt}) can be used to define the blocks in Figure \ref{fig:exx4} that go from $S_1^{(2)}$ to $(S_1^{(2)}S_2)^{(1)}$, from $S_2$ to $(S_1^{(2)}S_2)^{(2)}$,  from $Y_2$ to $(Y_2S_3)^{(1)}$ and from $S_3$ to $(Y_2S_3)^{(2)}$.    

As an example, the structure of matrices (\ref{eq:marg1}) and (\ref{eq:margt}) for $X_1$, $X_2$ and $X_3$ having sizes $M_1=2$, $M_2=3$ and $M_3=2$ respectively,  are 
{\footnotesize 
\begin{equation}
P(X_1|X_1X_2X_3)=
\left(\begin{array}{cccccc}
1   &  0 \\
1   &  0 \\
1   &  0 \\
1   &  0 \\
1   &  0 \\
1   &  0 \\
0   &  1 \\
0   &  1 \\
0   &  1 \\
0   &  1 \\
0   &  1 \\
0   &  1 
\end{array}\right);  
P(X_2|X_1X_2X_3)=
\left(\begin{array}{ccc}
1  &   0  &    0 \\
1  &   0  &    0 \\
0  &   1  &   0 \\
0  &   1  &   0 \\
0  &   0  &   1 \\
0  &   0  &   1 \\
1  &   0  &   0 \\
1  &   0  &   0 \\
0  &   1  &   0 \\
0 &    1  &   0 \\
0 &    0  &   1 \\
0 &    0  &   1 \\
\end{array}\right);  P(X_3|X_1X_2X_3)=
\left(\begin{array}{cccccc}
1   &  0 \\
0   &  1 \\
1   &  0 \\
0   &  1 \\
1   &  0 \\
0  &  1 \\
1   &  0 \\
0   &  1 \\
1  &  0 \\
0   &  1 \\
1  &  0 \\
0   &  1 
\end{array}\right);
\end{equation}
$\begin{array}{l}
P((X_1X_2X_3)^{(1)}|X_1)={1 \over 6}  P(X_1|X_1X_2X_3)^T ; \\
P((X_1X_2X_3)^{(2)}|X_2)={1 \over 4} P(X_2|X_1X_2X_3)^T ; \\
P((X_1X_2X_3)^{(3)}|X_3)={1 \over 6}  P(X_3|X_1X_2X_3)^T.
\end{array}$
}

\section*{Appendix B: Derivation of the ML Algorithm}
\label{app:ML}
To solve the ML problem (\ref{eq:mlp}) we seek to minimize the following cost function 
\begin{equation}
C(\theta)=-\sum_{n=1}^N L[n] \left[ \log({\bf f}_{X[n]}^T \theta {\bf b}_{Y[n]}) - {\bf 1}_{M_Y}^T \theta^T {\bf f}_{X[n]}\right],
\end{equation} 
where  we have added an extra term that when $\theta$ is row-stochastic and ${\bf f}_{X[n]}$ is a distribution, is just equal to one $\forall$ $n$.  This term avoids temporary divergence of $\theta$  and leads to a very stable algorithm.  
The constrained problem is: 
\begin{equation}
\left\{
\begin{array}{l}
\min_{\theta} C(\theta); \\
-\theta_{ij} \le 0, ~~~i=1:M_X;~j=1:M_Y; \\
\sum_{j=1}^{M_Y} \theta_{ij} -1=0,~~~i=1:M_X.
\end{array}
\right. 
\end{equation}
Note that the last two conditions automatically imply that $\theta_{ij} \le 1$. There are $M_X \times M_Y$ inequality  and $M_X$ equality constraints. Applying KKT conditions \citep{Avriel2003}, we get 
\begin{equation}
\left\{
\begin{array}{l}
{\partial C(\theta) \over \partial \theta_{lm}}+\sum_{i=1}^{M_X} \sum_{j=1}^{M_Y} \lambda_{ij} {\partial (-\theta_{ij}) \over \partial \theta_{lm}}+\sum_{i=1}^{M_X} \beta_i {\partial \over \partial \theta_{lm}}(\sum_{j=1}^{M_Y} \theta_{ij} -1 ) =0; \\
-\theta_{lj} \le 0;~ \sum_{j=1}^{M_Y} \theta_{lj} -1 =0; \\
\lambda_{lm} \ge 0;~\lambda_{lm}(- \theta_{lm})=0;
l=1:M_X; ~m=1: M_Y. 
\end{array}
\right.
\end{equation}
Taking the derivatives, 
\begin{equation}
\left\{
\begin{array}{l}
- \sum_{n=1}^N L[n] \left( { f_{X[n]}(l) b_{Y[n]}(m) \over {\bf f}_{X[n]}^T \theta {\bf b}_{Y[n]} } -  f_{X[n]}(l) \right) - \lambda_{lm}=0; \\
-\theta_{lj} \le 0;~ \sum_{j=1}^{M_Y} \theta_{lj} -1 =0; \\
\lambda_{lm} \ge 0;~\lambda_{lm}(- \theta_{lm})=0;
l=1:M_X; ~m=1: M_Y. 
\end{array}
\right. 
\end{equation}
Therefore $\lambda_{lm}$ and the complementary slackness conditions become 
\begin{equation}
\begin{array}{l}
\lambda_{lm} =- \sum_{n=1}^N L[n] \left( { f_{X[n]}(l) b_{Y[n]}(m) \over {\bf f}_{X[n]}^T \theta {\bf b}_{Y[n]} } -  f_{X[n]}(l) \right) ; \\
 \sum_{n=1}^N L[n] \left( \theta_{lm} { f_{X[n]}(l) b_{Y[n]}(m) \over {\bf f}_{X[n]}^T \theta {\bf b}_{Y[n]} } -  \theta_{lm}f_{X[n]}(l) \right)=0; 
l=1:M_X; ~m=1: M_Y. 
\end{array}
\end{equation}
It is easy to verify that $\lambda_{lm} \ge 0$ for all $l,m$.  From the complementary slackness condition and the constraints, the iterations for the ML algorithm immediately follow. To avoid occasional divisions by zero it is advised to run the algorithm keeping all the distribution elements greater than zero. In case of instantiated variables a very small value can replace all the null values.   

\vskip 0.2in
\bibliography{ProbPropBib}

\end{document}